\documentclass{article}

\usepackage{todonotes}

\usepackage[preprint, nonatbib]{neurips_2019}

\usepackage{wrapfig}
\usepackage[utf8]{inputenc} % allow utf-8 input
\usepackage[T1]{fontenc}    % use 8-bit T1 fonts
\usepackage{hyperref}       % hyperlinks
\usepackage{url}            % simple URL typesetting
\usepackage{booktabs}       % professional-quality tables
\usepackage{amsfonts}       % blackboard math symbols
\usepackage{amsmath}
\usepackage{amssymb}
\usepackage{amsthm}

\usepackage{xcolor}
\usepackage{float}
\usepackage{subfig}
\usepackage{algorithm}
\usepackage[noend]{algpseudocode}
\usepackage{graphicx}
\usepackage{graphics}
\usepackage{nicefrac}       % compact symbols for 1/2, etc.
\usepackage{microtype}      % microtypography
\usepackage{wrapfig,lipsum,booktabs}
\usepackage{placeins}
\usepackage{multirow}

\title{Hierarchical Symbolic Reasoning in Hyperbolic Space for Deep Discriminative Models}

\author{%
  Ainkaran Santhirasekaram* $\dagger$\\
  \texttt{a.santhirasekaram19@ic.ac.uk} \\
  \And
  Avinash Kori* $\dagger$\\
  \texttt{a.kori21@ic.ac.uk} \\
  \And
  Andrea Rockall $\ddagger$\\
  \texttt{a.rockall@ic.ac.uk} \\
  \AND
  Mathias Winkler $\ddagger$\\
  \texttt{m.winkler@ic.ac.uk} \\
  \And 
  Francesca Toni $\dagger$\\
  \texttt{f.toni@ic.ac.uk} \\
  \And
  Ben Glocker $\dagger$\\
  \texttt{b.glocker@ic.ac.uk} \\
  \AND
  * joint first author \\
  $\dagger$ {Department of Computing, Imperial College London}\\
  $\ddagger$ {Department of Surgery and Cancer, Imperial College London}
  }
\begin{document}

\maketitle

\begin{abstract}
Explanations for \emph{black-box} models help us understand model decisions as well as provide information on model biases and inconsistencies. %which makes the explainability arguably a promising direction of research.  
Most of the current explainability techniques provide a single level of explanation, often in terms %focus on proving 
of feature importance scores or feature attention maps in input space.
Our focus is on  
explaining deep discriminative %learning 
models at \emph{multiple levels of abstraction}, from fine-grained to fully abstract explanations.
We achieve this by using the natural properties of \emph{hyperbolic geometry} to more efficiently model a hierarchy of symbolic features and generate \emph{hierarchical symbolic rules} as part of our explanations.
Specifically, for any given %classifier model
deep discriminative model, we distill the underpinning knowledge by discretisation of the continuous latent space using vector quantisation to form symbols, followed by a \emph{hyperbolic reasoning block} to induce an \emph{abstraction tree}.
%As a part of explanations, 
We traverse the tree to extract
explanations in terms of symbolic rules and its corresponding visual semantics.
We demonstrate the effectiveness of our method on the MNIST and AFHQ high-resolution animal faces dataset.
Our framework is available at \url{https://github.com/koriavinash1/SymbolicInterpretability}.
\end{abstract}

\section{Introduction}
\label{section:intro}
% Explanations for a \emph{'black-box'} classifier come in many different forms; recently, there has been a dramatic increase in explainable AI (XAI) approaches.
Explainable AI (XAI) aims to improve the transparency and trustworthiness of deep learning models \cite{doshi2017role, kroll2015accountable}; XAI can help with identifying biases, which is important for the safe and fair use of prediction models \cite{kim2018interpretability}.
XAI approaches can be broadly categorized into ante-hoc and post-hoc methods \cite{lipton2018mythos}. 
% Explaining decisions made by classifiers can not only help us understand the mechanism followed by the model but also uncover model biases \cite{kim2018interpretability}, which helps in better understanding the data-generating process \cite{narayanaswamy2020scientific}.
Ante-hoc explainability focuses on developing inherently transparent models. For example,  \cite{rightforright} introduces a symbolic reasoning block to promote transparency during training while 
\cite{tellmewhy} uses natural language descriptions that complement model predictions.
\cite{debate} develops intrinsically aligned models via a debating game with a human judge.

Post-hoc explanations are the most commonly explored approaches, including explanations via feature-attribution \cite{lime, SHAP, deeplift}, saliency maps \cite{selvaraju2017grad, gradcampp, integratedcam} counterfactuals \cite{goyal2019counterfactual} or concept extraction \cite{tace, dissect, concepttrail}. Feature attribution methods \cite{lime, SHAP} focus on assigning importance weighting to each and every feature in an input space, indicating their contribution towards the classifier's decision; these methods are also considered shallow example-based explanations as they do not provide any further feedback %to
on the model. 
Saliency-based methods \cite{selvaraju2017grad} 
generate attention maps in an input space indicating %the image 
regions responsible for deriving the classifier's decision. These methods %are 
also %known as 
provide
example-based explanations. However, both feature-attribution and saliency-based explanations do not highlight the model's perceived feature interactions as %it 
they only work %s 
on input data, which makes  explanations a function of the data as well as of the model. This makes it difficult to assess the faithfulness of explanations towards the model's reasoning.
% \cite{bengio2013representation} discuss, broadly, the properties of effective feature representations and feature disentanglement, which is a building block of concept-based explanations. 
% Methods like \cite{tace} construct global level features but do not capture interactions between them, while \cite{concepttrail, dissect} do consider both existences of features and their interactions.
% Both these methods work on a continuous domain and when the feature behavior is non-monotonic, the stability and coherence of the obtained explanations are questionable. Also, most of these concept-based explanations are a function of both data and model, which makes it hard to assess the faithfulness of explanations towards the model's reasoning.
% Concept-based explanations aim to learn a transformation that helps in representing input data with a relatively-smaller number of latent factors with useful orthogonal properties \cite{locatello2019challenging, adel2018discovering, creager2019flexibly}.
Counterfactual explanation methods \cite{sauer2021counterfactual, chang2018explaining, nemirovsky2020countergan, lang2021explaining} help in analysing a classifier by creating several carefully constructed \emph{what-if} scenarios by perturbing specific features, but are also example-based.
% The effectiveness of counterfactual explanations solely depends on an ability to tweak features to construct desired \emph{what-if} scenarios. 

% Learning effective representation doesn't only help in improving performance but also improves generalizability and model biases \cite{bengio2013representation}.
% \cite{higgins2021unsupervised} describes that representations learned by macaque inferotemporal cortex closely resembles disentangled representation in $\beta-$VAE\cite{higgins2016beta}, claiming existence of concepts based reasoning in brains. Additionally, 

These methods overall provide single stage explanations but do not consider concept-based reasoning as perceived by humans \cite{armstrong1983some, burnston2021evolving}. Concept-based explanations \cite{tace, dissect, concepttrail} go beyond feature-attribution and saliency-based methods by constructing higher-level concepts indicating their influence on the classifier's decision.
\cite{concepttrail, dissect} focus %es 
on developing traces, which indicate the flow of reasoning to make certain predictions, %and 
thereby considering feature interaction. Th%is approach is
ese approaches are closely related to a %new FT: not new at all - see e.g. https://dblp.org/pid/26/4820.html
popular 
wave of AI termed neuro-symbolic learning \cite{hudson2019learning, vedantam2019probabilistic, mao2019neuro, yi2018neural}, where a data-driven  deep learning method 
%such as deep learning 
is used to learn sub-symbolic representations to denote concepts while exploiting symbolic methods to capture reasoning.  
In fact, studies in computational neuroscience have claimed that concept representation in the brain can be in a finite discrete form rather than using continuous and infinitely many representations \cite{tee2020information}.
% Psychological studies have also shown that the key aspect for consistent feature representation is the ability to build a visual hierarchy to create multiple distinguishable manifold representations \cite{liao2016learning}. 

The notion of hierarchical concept-based reasoning is the most commonly posited learning principle in systems neuroscience \cite{burnston2021evolving, vetter2014varieties, wessinger2001hierarchical, meunier2009hierarchical}. 
This form of reasoning has already been employed in neuro-symbolic AI \cite{glanois2021neuro, mitchener2022detect} to develop hierarchical representations in the form of knowledge/abstractions graphs and induce hierarchical rules as explanations. However these approaches are limited in developing hierarchical visual reasoning for classification tasks due to a continuous latent space. They do not also consider the natural geometry of the space upon which to develop hierarchical or neuro-symbolic models. 
% However, it is unclear how the brain acquires hierarchical representations or how conceptual knowledge can be formulated at different levels of abstractions or hierarchical organization. 
% Multiple neuroscience studies have shown that humans rely on consistent feature representation irrespective of innate variations of objects \cite{karimi2017invariant, nielsen2008object}. 
% Inspired by this form of reasoning we proposed hierarchical reasoning in hyperbolic space to explain \emph{blackbox} classifiers.

Inspired by hierarchical reasoning, we tackle the aforementioned issues to develop hierarchical symbolic explanations for deep discriminative models trained on imaging data. Our approach distills the knowledge from a trained classifier into a discrete surrogate model. We consider the natural geometry in which to develop our discrete surrogate model and form a hierarchical symbolic  representation in hyperbolic space of the visual world with the learned interactions between symbols forming an abstraction tree \cite{sharma2005probabilistic}.
This method makes generated explanations faithful to the classifier rather than the data. It also addresses the issue of non-monotonic behavior of features, by considering discrete symbols which can either be sampled or not.
Our main contributions in this work include:

\begin{itemize}
    \item \textbf{Symbol formation (Sections~\ref{subsec:symbolformation}, \ref{subsec:reasoning}):} A method to discretise the continuous latent space of a given classifier model into a hierarchy of discrete vectors denoted as symbols.  We exploit the natural structure of hyperbolic geometry to more efficiently model this hierarchy. 
    % preserve hierarchy and semantic distances between symbols by  modelling the hierarchy in hyperbolic space.
    \item \textbf{Symbol abstraction and hyperbolic reasoning module (Section~\ref{subsec:reasoning}):} An effective way of learning symbol conjunction using binary weight layers to form an abstraction tree, which %will
    can
    provide both local image-level and global class-level hierarchical rules.
    \item \textbf{Explanations (Section~\ref{subsec:explanations}):} A way of obtaining visual semantics for any given symbol in a hierarchy.
\end{itemize}
% To facilitate future work, all code will be released upon publication.

\section{Background}
\label{section:bg}

\textbf{Vector quantisation: }
\cite{van2017neural} proposed a new way of training variational autoencoders (VAE) with discrete latent variables, which showed competitive performance with its continuous counterparts. 
They achieve the discretisation of the continuous latent variables from the output of an encoder with a method called vector quantisation (VQ). 
% Here, the posterior and prior distributions are categorical and are sampled from an embedding table called codebook by indexing the embedding vector by nearest neighbour look up, which serves as an input to the decoder. 
% However, training this codebook is not straightforward as one cannot backpropogate the gradients through a sampling based operation to circumvent this problem, the author proposed a gradient approximation whereby the gradients from the decoder and copied over to the encoder.
% Alternative methods to learn a discrete latent space include using a Gumbel-Softmax distribution \cite{jang2016categorical}.
% % a continuous distribution with a temperature parameter that is constantly annealed during training to converge to a discrete distribution \cite{jang2016categorical}. 
% However, at the end of the training, the gradients in the Gumbel-Softmax approach are noted to have a larger variance than in the case of VQ \cite{van2017neural}. 
We therefore use VQ to learn a set of discrete symbols to represent the continuous latent space of a classifier.

\textbf{Hyperbolic embeddings: }
A natural objective when embedding symbolic data in graphs is for the distances between symbols, defined by the space which they reside in, to correlate with their semantic similarity. 
However, to model increasingly complex relations between symbols, one is bounded by the dimensionality of embeddings in Euclidean space \cite{nickel2014reducing}. 
% For example, Nickel et al. \cite{nickel2014reducing} demonstrated the difficulties to embed data in a graph without loss of information. 
% We can define the graph volume as the number nodes within an arbitrary radius from the center node. 
This is because the number of nodes in most cases grows exponentially as the graph distance from the centre node increases, while Euclidean space grows polynomially. This leads to embedding distortion and loss of information \cite{sala2018representation}. 
Hyperbolic geometry is a form of non-Euclidean geometry with a constant negative Gaussian curvature whose space grows exponentially. One can even informally describe hyperbolic space as the continuous version of trees, making it naturally equipped to deal with tree-like structures.
This property has therefore been exploited in the literature for embedding hierarchical data in hyperbolic space without distortion and provides the reasoning for embedding our abstraction tree in hyperbolic space. 
% , addressing the issue of loss of information in Euclidean embeddings. 
For example, in \cite{nickel2017poincare}, the authors proposed to use a hyperbolic model called the Poincare disk for word embeddings to simultaneously capture hierarchy and similarity reliably, which led to promising results in link prediction. 
% They demonstrated the structural bias imposed in hyperbolic embedding produces superior performance to their euclidean counterparts in both link prediction and reconstruction to evaluate representation capacity.
Hyperbolic neural networks were proposed to perform all the operations of a neural network in hyperbolic space \cite{ganea2018hyperbolic}. 
This was later exploited in the development of hyperbolic graph convolutional neural networks (HGCNN) \cite{chami2019hyperbolic, liu2019hyperbolic} 
which forms the structure of our  method for reasoning in hyperbolic space. 

\textbf{Symbolic reasoning: } 
% \cite{hudson2019learning, vedantam2019probabilistic, mao2019neuro, yi2018neural} propose neuro-symbolic learning approaches where a data-driven method is used to learn sub-symbol representations while symbolic methods being used in reasoning aspects.
Reasoning with symbolic data can mainly be performed via inductive or deductive approaches.
%FT I would argue that abductive reasoning is as, if not more, important - we can discuss at some point
Inductive reasoning considers a specific dataset and makes a broad generalization that is most probable while deductive reasoning provides complete evidence of the truth for its conclusion.
Inductive logic programs (ILP) is a framework which learns using relational theories, such as using heuristics and physical properties to understand images \cite{muggleton2018meta}.
Due to the strong inductive biases imposed by background knowledge, ILP approaches usually need very few examples to learn and generalize well \cite{lin2014bias}. 
% However, at the same time this robust approach makes it difficult for ILP methods to fit well to noisy real world datasets (no reference).
In ILP, all the known hypotheses of an environment and a model's reasoning can be converted to logic programs that can be read and understood by humans. This makes ILP an ideal candidate for explainable AI \cite{michie1988machine} %which 
and 
forms the basis for learning our abstraction tree. 
% In our work, we  extract local and global hierarchical rules from an induced abstraction tree.  
% Recently \cite{payani2019inductive} proposed a differentiable deep logic network that can learn and represent boolean functions explicitly to solve ILP problems.

\section{Methods} 

\begin{figure}[]
    \centering
    \includegraphics[width=1\textwidth]{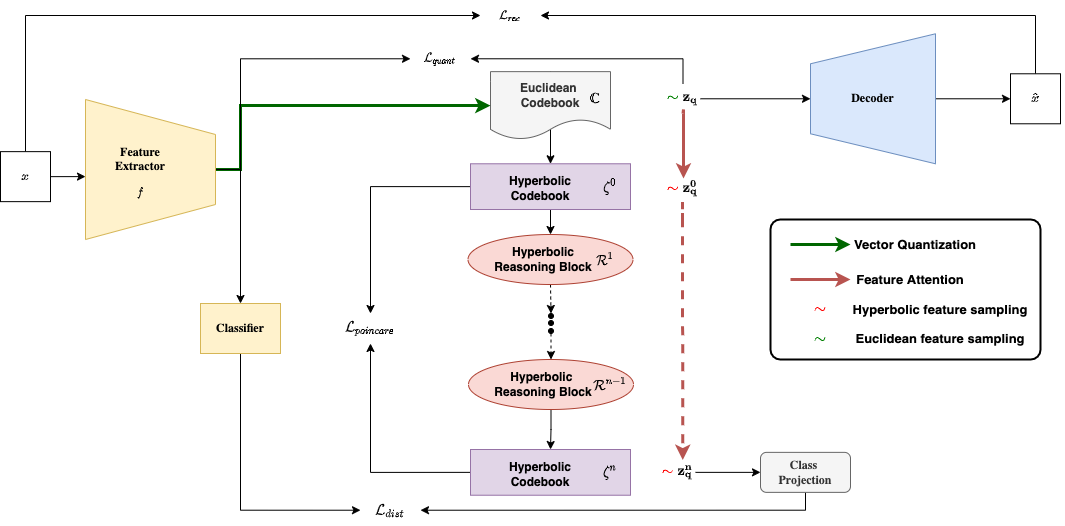}
    \caption{Overview of the proposed framework, in which the feature extractor and classifier %components 
    describe the trained blocks of the given model. 
    The euclidean codebook forms a discrete representation of the continuous latent space from a feature extractor followed by hyperbolic codebooks and reasoning blocks to obtain a hierarchy of abstractions, %that
    which is later used in extracting explanations. 
    The decoder block is independently trained to obtain visual semantics for %an
    extracted symbolic rules.}
    \label{fig:overview}
\end{figure}

Our framework consists of three steps: (i) Discretisation of a classifier's latent space for symbol formation; (ii) Generating a hierarchy of symbols in hyperbolic space to construct an abstraction tree; and (iii) Rule/explanation extraction.
Figure \ref{fig:overview}, %illustrates 
overviews
our overall proposed framework.

\subsection{Preliminaries and notations}
\label{subsection:prelims_notations}

% \begin{asu} 
% \label{assumption:classifier_decomposition}
% In this work, we only consider classifiers of the form $\mathcal{C} = f \circ g$, where feature extractor $f:\mathcal{X} \rightarrow \mathcal{E}$ maps input images to latent vector $\mathcal{E} \in \mathbb{R}^l$ and feature classifier $g:\mathcal{E} \rightarrow \mathcal{Y}$ maps latent space to class labels.
% \end{asu}
\subsubsection{Hyperbolic geometry:}

First, we introduce some important geometric concepts.
A $n$ dimensional \textbf{Manifold}  $\mathcal{M}$ is a topological space that can be locally approximated in Euclidean space $\mathbb{R}^n$.
% , that is each point on the manifold consists of a neighbourhood homeomorphic to an open subset in $\mathbb{R}^n$. 
The \textbf{Tangent Space} $\mathcal{T}_x\mathcal{M}$ is defined as an $m$ dimensional vector space which is a first order approximation of a point $x$ on $\mathcal{M}$. 
The \textbf{Riemannian Metric} defines the set of inner products $g_x: \mathcal{T}_x\mathcal{M} \times \mathcal{T}_x\mathcal{M} \rightarrow \mathbb{R}$ of every point $x$ on $\mathcal{M}$.
% and a \textbf{Riemannian Manifold} defines a smooth manifold with a Riemannian metric. 
\textbf{Parallel transport} $P_{x \rightarrow y}$  describes the translation of a vector field $V$ along a differentiable manifold to a new vector field $V'$ such that the covariant derivative always stays at 0.
% i.e. $V'$ is parallel to $V$.

We now introduce the \textbf{Hyperboloid} and \textbf{Poincare} models of hyperbolic space equipped with constant negative curvature $-1 /K, (K > 0)$.
% which is inversely proportional to the radius of the Hyperbolic model. 
In the rest of this works, we work with only hyperboloid $\mathbb{H}^{d, K}$ and Poincare $\mathbb{B}^{d, K}$ models of $d$ dimensions with unit radius and hence fixed negative curvature of -1; $\mathbb{H}^{d, 1}$, $\mathbb{B}^{d, 1}$. 
We focus on the two sheet unit hyperboloid model defined by Riemmanian metric  $g^{\mathbb{H}, 1}_x$ given by the Minkowski metric tensor $\langle ., .\rangle _S$ whereby $\langle x, x \rangle_S = -1$ and $x \in \mathbb{R}^{d+1}$ \cite{chami2019hyperbolic}. 
Therefore, the dot product between arbitrary points $u$ and $v$ on $\mathbb{H}^{d,1}$, is defined as   $g^{\mathbb{H}, 1}_x\langle u, v \rangle :=  - u_0 v_0 + u_1 v_1 \ldots + u_d v_d;  u_0 > 0$ \cite{chami2019hyperbolic}. Given the definition of the Riemannian metric tensor, we can introduce the notion of the geodesic distance defined as the shortest distance between $(u,v)\in \mathbb{H}^{d, 1}$. 
% In \cite{chami2019hyperbolic}, we note the derivation of  $\mathbb{H}^{d, 1}$ for defining 
Geodesic distance in the hyperboloid is given as in \cite{chami2019hyperbolic}:
$d^{\mathbb{H}, 1} (u,v) = arcosh(g^{\mathbb{H}, 1}_x\langle u, v \rangle)$ .

We next formally define the Poincare unit ball, $\mathbb{B}^{d, 1} = \{ x \in \mathbb{R}^{d} \vert \lVert x \rVert < 1$\}. 
Geodesics in $\mathbb{B}^{d, 1}$ are arcs of circles orthogonal to the boundary. $\mathbb{B}^{d, 1}$ is determined by the following Riemmanian metric tensor: $g^{\mathbb{B}, 1}_{x}  =  \Big(\frac{2}{1 - \lVert x \rVert ^2}\Big)^2$.
This allows to define the geodesic distance for $(u,v)\in \mathbb{B}^{d, 1}$ below:
%\ref{eqn:pd}
\begin{equation}
    d^{\mathbb{B}, 1} (u,v) \!=\!  1 + 2 \Big(\frac{\lVert u - v \rVert ^ 2}{(1 - \lVert u \rVert ^2)(1 - \lVert v \rVert ^2)}\Big) 
    \label{eqn:pd}
\end{equation}

Now we introduce the notion of mapping between hyperbolic and euclidean space (tangential plane). 
Given $y \in \mathcal{T}_x\mathbb{H}^{d, 1}$ and $v \in \mathbb{H}^{d, 1}$,  
exponential and logarithmic  mappings are denoted %such that 
$exp_{x}^{\mathbb{H}, 1}(y)$: $\mathcal{T}_x\mathbb{H}^{d, 1} \rightarrow \mathbb{H}^{d, 1}$ and $log_{x}^{\mathbb{H}, 1}(v)$`: $\mathbb{H}^{d, 1} \rightarrow \mathcal{T}_x\mathbb{H}^{d, 1}$ respectively. 
In this work, we perform mappings with the tangential space at the origin $o$, which for $\mathbb{H}^{d, 1}$ is defined on $\{1, 0_1, \ldots, 0_d\} \in \mathbb{H}^{d, 1}$. 
In this case, the mappings between $y \in \mathcal{T}_o\mathbb{H}^{d, 1}$ and $v \in \mathbb{H}^{d, 1}$ such that $\langle y, o \rangle = 0$  are calculated as follows: %\ref{eqn:maph}
\begin{equation}
    exp_{o}^{\mathbb{H}, 1}(y) \!=\! \bigg(cosh(\lVert y_{1:d} \rVert _2), sinh(\lVert y_{1:d} \rVert _2)\frac{y_{1:d}}{\lVert y_{1:d} \rVert _2}\bigg); %\quad
 log_{o}^{\mathbb{H}, 1}(v) \!=\!  \bigg(0, arcosh(v_0) \frac{v_{1:d}}{\lVert v_{1:d} \rVert _2}\bigg)
    \label{eqn:maph}
\end{equation}

Next, the equations for the mappings between $y$ on $\mathcal{T}_o\mathbb{B}^{d, 1}$ and $v$ on $\mathbb{B}^{d, 1}$, with $o$ defined on $\{0, 0_1, \ldots, 0_d\} \in \mathbb{B}^{d, 1}$, is shown below: %\ref{eqn:pmt}.
\begin{equation}
    exp_{o}^{\mathbb{B},1}(y) = \bigg(tanh(\lVert y \rVert _2)\frac{y}{\lVert y \rVert _2}\bigg); \quad log_{o}^{\mathbb{B},1}(v) =  \bigg(0, arctanh(\lVert v \rVert _2) \frac{v}{\lVert v \rVert _2}\bigg)
    \label{eqn:pmt}
\end{equation}

We apply projections as described in \cite{chami2019hyperbolic} to constrain points to the manifolds during optimisation. Please refer to appendix \ref{sec:appendix} for formal definitions of projection. 

Next, $\mathbb{H}^{d, 1}$ and $\mathbb{B}^{d, 1}$ are both isomorphic and therefore there exists a diffeomorphism mapping $\psi$ between $u \in \mathbb{H}^{d, 1}$ and $v \in \mathbb{B}^{d, 1}$ as highlighted %in equation \ref{eqn:hp}.
below:
\begin{equation}
    \psi_{\mathbb{H}^{d, 1}\rightarrow\mathbb{B}^{d, 1}}(u_0...u_d) = \frac{u_1...u_d}{u_0 +1}; \quad \psi_{\mathbb{B}^{d, 1} \rightarrow \mathbb{H}^{d, 1}}(v_1...v_d) = \frac{(1 + \lvert \lvert v \rvert \rvert _{2}^{2}, 2v_1...2v_d)}{1 - \lvert \lvert v \rvert \rvert_{2}^{2}}
    \label{eqn:hp}
\end{equation}

In our work, we apply feature transformations in hyperbolic space, %which uses
using 
hyperbolic linear layers \cite{ganea2018hyperbolic,chami2019hyperbolic}. 
The operations of Mobius addition $\oplus$ and Mobius scalar multiplication $\otimes$ in hyperbolic space can be shown to be analogous to the euclidean vector space operations of scalar multiplication and addition.
\cite{ganea2018hyperbolic} proves that Mobius scalar multiplication is equivalent to applying a logarithmic mapping of point $v \in \mathbb{B}^{d, 1}$ to the tangential space at $o$ and multiplying the scalar $r$ before mapping the scaled point back to Poincare space highlighted % in \ref{eqn:mt}. 
below:
\begin{equation}
     r \otimes^1 v  = exp_{o}^{\mathbb{B}, 1} (r \ log_{o}^{\mathbb{B}, 1}(v))
    \label{eqn:mt}
\end{equation}

In a hyperbolic linear layer, we also add a bias term $b$. \cite{ganea2018hyperbolic} derives a simple equivalent solution to Mobius addition ($x \oplus b$) shown in Equation~\ref{eqn:mp}. One firstly defines $b$ on $\mathcal{T}_o\mathbb{B}^{1,d}$ which is parallel transported to the tangential space $\mathcal{T}_x\mathbb{B}^{1,d}$ before mapping back to Poincare space. Please refer to appendix \ref{sec:appendix} for the equations of parallel transport as well as the general form for $exp_{x}^{K}(y)$ and $log_{x}^{K}(v)$.
\begin{equation}
    v \oplus ^1 b  = exp_{v}^{\mathbb{B}, 1} (P_{o \rightarrow v}^1(b))
    \label{eqn:mp}
\end{equation}

Equations \ref{eqn:mt} and \ref{eqn:mp} can also apply to scalar multiplication and addition in $\mathbb{H}^{d, 1}$ \cite{chami2019hyperbolic}. 
We now let $W \in \mathbb{R}^{d' \times d}$ and $B \in \mathbb{R}^{d}$ to help define the hyperbolic linear feature transformation $h(x)$ in $\mathbb{H}^{d, 1}$ and $\mathbb{B}^{d, 1}$ by combining Equations \ref{eqn:mt} and \ref{eqn:mp} to form: %equation \ref{eqn:ht}. 

% ; $h(x) = (w \otimes^1 x) \oplus^1 b$.
\begin{equation}
    h(x) = (W \otimes^1 x) \oplus^1 B
    \label{eqn:ht}
\end{equation}

\subsubsection{Model discretisation}

%Let $(\mathcal{X},\mathcal{Y})$ belong to dataset $\mathcal{D}$, such that $\mathcal{X} \in \mathbb{R}^{p \times p}$ and $\mathcal{Y} = \{1, 2, 3, ..., N\}$, where $p \times p$ corresponds to the image dimension and $N$ corresponds to total number of classes in a given dataset.
Let $\mathcal{D} \subseteq \mathcal{X} \times \mathcal{Y}$ be the dataset, such that elements of $\mathcal{X}$ are in  $\mathbb{R}^{p \times p}$ and $\mathcal{Y} %\in 
= \{1, \ldots, N\}$, where $p \times p$ and $N$ correspond to the dimension of an input image and total number of classes, respectively.
In this works, we only consider classifiers of the form $\mathcal{C} = f \circ g$, where feature extractor $f:\mathcal{X} \rightarrow \mathcal{E}$ maps input images to the latent space $\mathcal{E}$ and the feature classifier $g:\mathcal{E} \rightarrow \mathcal{Y}$ maps the latent space to class labels. 
We aim to distill the knowledge of a continuous classifier into a hierarchy of related symbols to form an abstraction tree whereby a symbol is a discrete concept represented as a vector. This is based on the valid assumption that the visual world consists of discrete concepts which are related in a hierarchical manner such that abstract visual concepts are the amalgamation of fine-grained discrete features.
%TODO remark
% This assumption is valid to an extent, as in most of the natural image concepts are formed by a large number fine-grained features.
Therefore, given that the abstraction tree corresponds to an internal representation of the visual world that grows exponentially, this provides the basis to better embed the tree with minimal distortion in hyperbolic $\mathcal{H}$ rather than Euclidean space $\mathbb{R}$ \cite{nickel2017poincare}. 
The $i^{th}$ level in the tree is represented as a hyperbolic codebook $\zeta^i \in \mathcal{H}^{M_i \times d}$ which provide different levels of abstraction for the continuous space $\mathcal{E}$, as %illustrated 
shown
in Figure \ref{fig:overview}. 
The total number of codebooks $n$ ($\zeta^n$) and embedding dimensionality $d$ are hyper-parameters that are selected based on the use case. 
$M_0, M_1, \ldots, M_n$ are hyper-parameters corresponding to the total number of vectors in codebooks $\zeta^0, \zeta^1 \ldots , \zeta^n$ respectively. 
The greater $n$, the more levels of abstraction %, which therefore provide 
and the 'deeper' %level 
the explanations. 
However, we limit $n$ to an appropriate number to prevent the exponential increase in rules as we strive for an Occam's Razor approach to generate explanations. 
Given this approach, we carry out ablations to find the smallest $M_n$ in each codebook to achieve a minimum of 90\% knowledge distillation from a pre-trained classifier.

Formally, an abstraction tree is developed by learning the function $\mathcal{K}$ which collapses the Euclidean continuous latent space $\mathcal{E}$ into ideally $\lfloor log_2 N + 1\rfloor$ symbols in $\zeta^n$ as this represents the minimum number of positive symbols to encode $N$ classes. We can decompose $\mathcal{K}$ such that $\mathcal{K} = \mathcal{R} \circ \mathcal{VQ} \circ \mathcal{MD}$, where $\mathcal{MD}$ denotes a feature modulation layer before discretisation by vector quantisation  ($\mathcal{VQ}$) and $\mathcal{R}$ expresses the hyperbolic reasoning module. $\mathcal{R}$ can further be decomposed into $\mathcal{R}^n \circ \mathcal{R}^{n-1} \circ \ldots \circ \mathcal{R}^{1}$ with the output of each $\mathcal{R}^l$ computing a different level of abstraction in the form of $\zeta^i$ %visualised in 
(see again Figure \ref{fig:overview}). We train $\mathcal{K}$ by sampling $z$ from $\mathcal{E}$ and sequentially mapping and discretising $z$ to increasing levels of abstraction in the abstraction tree to produce $z_{q}^{i}$ with the final level of abstraction used to classify $z$.

% For classifying/distilling knowledge from pretrained classifier we define quantised classifier $q$ which maps quantised symbols from last quantisation block $\zeta^n$ to class labels $\mathcal{Y}$. 
% For visualising the effect of each symbol we define decoder model $\mathbb{D}$ which maps quantized symbols from first quantisation block $\zeta^1$ to image space $\mathcal{X}$.

\subsubsection{Inductive logic programming} 
Logic programming is a %type of programming language FT: it is not - Prolog is, but it has many non-logical features
knowledge representation and reasoning formalism which helps us %to 
describe %the 
relations  of interest in terms of facts and rules expressed in %using 
formal logic.
Here, %the relations 
rules are usually described in the form of clauses 
as described in Equation~\ref{eqn:example}, where $H$ is the \emph{head} of the clause and $B_1, %B_2, B_3, ...,
\ldots, B_n$ is the \emph{body} of the clause. 
\begin{equation}
    H \leftarrow B_1, %B_2, B_3,
    \ldots ,B_n \quad (n \geq 1)
    \label{eqn:example}
\end{equation}
A clause of this form captures relational information of the form: the head is true when all the elements in the body are true. 
In this work, we assume that $H$ and each and every %term in  FT: term means something very specific in logic, it is a constant, a var or a fucntor applied to a term, you do not wnat to use term here
$B_i$ are atoms, which are Boolean functions %called predicates of some variables
amounting to applying predicates to terms (constant or variables).

Inductive Logic Programming (ILP) is a form of logic programming %that helps us 
for learning these clauses %/hypothesis 
(as shown in Equation \ref{eqn:example}) 
from data.
ILP can be formally defined in terms of a tuple  $(\mathcal{B}, \mathcal{P}, \mathcal{N})$, where $\mathcal{B}, \mathcal{P}, \mathcal{N}$ correspond to background knowledge, positive and negative examples respectively.
% and are sets of clauses, atoms that need to be entailed, and atoms that should not be entailed when learning. 
The main objective for ILP is to formulate hypotheses (i.e. sets of clauses) given $\mathcal{B}, \mathcal{P},$ and $\mathcal{N}$.
These hypotheses are sets $\mathcal{C}$ of clauses of the form considered before% are usually expressed as a set of definite clauses $\mathcal{C}$
, such that $\mathcal{C}$ explains all the examples, formally described as $\mathcal{B}, \mathcal{C} \models e$ for all positive examples
 $e \in \mathcal{P}$ and $\mathcal{B}, \mathcal{C} \not \models e$  %\quad \forall 
 for all negative examples $e \in \mathcal{N}$. 
In general, ILP systems can be %solved with 
built  using two main approaches:
(i) Bottom-up, %this method starts 
starting with analysing $\mathcal{B}$ to construct a hypothesis that generalises well on $\mathcal{P}$ and $\mathcal{N}$, e.g., Progol \cite{muggleton1995inverse} 
%FT: wrong reference - Prolog is just for reasoning, not for learning FROM THE TOP OF MY HEAD I CANNOT THINK OF A REFERENCE - I WOULD ASK E.G. ALEX SPIES HOPING THAT HE CAN HELP WITH A REFERENCE FOR THE BOTTOM-UP APPROACH: AK: Fixed it. 
;
(ii) Top-down approach, where clauses are generated via some pre-defined templates and these clauses are tested for satisfiability on $\mathcal{P}$ and $\mathcal{N}$, e.g., as in Metagol \cite{metagol} and dILP \cite{evans2018learning}.
% Our approach falls under the former category, as we learn the relationship between different levels of abstractions using knowledge distillation via gradient descent.

TILDE \cite{blockeel1998top} formulates an approach to learn  ILP hypotheses via a first-order logical framework for top-down induction of logical decision trees.
First order logical decision trees are binary decision trees, where each node corresponds to a conjunction of atoms. % \cite{blockeel1998top} shows that logical decision trees are more expressive than a program induced by a traditional ILP system. FT: THIS IS OLD WORK, I DO NOT AGREE WITH THIS - AND YOU DO NOT NEED TO ALIENATE ANY READER, JUST DROP THIS SENTENCE
In this work, we build on the idea of binary trees to form the component of our hyperbolic reasoning blocks which learns via a bottom-up approach the hierarchical relations/conjunction of atoms in between the codebooks to develop our abstraction tree. We then use this tree to derive hierarchical rules. 
% Here, decision trees correspond to a relationship between different level of hierarchy in reasoning, and nodes in decision tree correspond to a conjunction of atoms from a previous layer in a hierarchy.

\subsection{Symbol formation}
\label{subsec:symbolformation}
The first step in our framework, symbol formation, initially consists of learning a feature modulation layer ($\mathcal{MD}$) which consists of batch normalisation followed by a $1 \times 1$ convolutional layer such that $z_m = \mathcal{MD}(z)$. This is followed by learning discrete symbols in the form of $d'$ dimensional vectors using vector quantisation ($\mathcal{VQ}$) to form a fixed sized  Euclidean codebook $\mathbb{C} \in \mathbb{R}^{M_0 \times d'}$ (Figure \ref{fig:overview}). We do not perform $\mathcal{VQ}$ in hyperbolic space as we found this significantly less stable. 
 
% We define a uniform discrete prior and learn the categorical distribution $\mathbb{P}(z_{q} \vert z)$ to learn the set of $d$ dimensional symbols which form  $\mathbb{C}$. 

Given $(z_q, z_m) \in \mathbb{R}^{K\times d'}$ and $k \in K$, we define a deterministic process which maps each embedding vector $z_{mk} \in z_m$ to the nearest Euclidean codebook vector to form $z_{qk} \in z_q$ shown %in equation \ref{eq:vq}.
below:
\begin{equation}
  \label{eq:vq}
    z_{qk} = argmin_j\|z_{mk} - \mathbb{C}_j\|_2,     \forall \quad k\in K
\end{equation}

% \begin{equation}
%   \label{eq:vq}
%     z_q = \mathbb{E}_j = \left\{
%     \begin{array}{ll}
%       1 \qquad   \text{for} \quad r = argmin_j\|z_{m} - \mathbb{E}_j\|_2\\
%       0 \qquad   \text{otherwise}
%     \end{array}
%     \right\}
% \end{equation}

Equation \ref{eq:vq} defines a sampling process which is non-differentiable but in order to update/learn the weights of $\mathcal{MD}$ and the symbols which form $\mathbb{C}$ based on this sampling method, we apply straight through 
gradient approximation. This then allows our discrete surrogate model to be trained end to end with the following Quantisation loss  \cite{van2017neural}: $\mathcal{L}_{quant} = \|sg(e)-\mathbb{C}\|_2 + \beta\|e-sg(\mathbb{C})\|_2$. 
We apply stop gradients (sg) to constrain updates to the appropriate operands \cite{van2017neural}. 
% such that only $\mathbb{E}$ is updated in the first loss term and $e$ is only updated in the second loss term, which is defined as the commitment loss. 
Our ablations determined an optimum value of 0.2 for $\beta$.
Next, we apply a linear layer with weights $w_e \in \mathbb{R}^{d \times d'}$ to the Euclidean codebook to reduce the dimensionality to the desired embedding dimensionality $d$ for $\zeta^i$ before applying an exponential mapping \ref{eqn:pmt} to Poincare space: $\zeta^0 \in \mathbb{B}^{M_0 \times d,1}$. We choose Poincare space in this work due to the enhanced visual interpretability of 2D embeddings on the Poincare disc \cite{nickel2017poincare}. We however note it may be more beneficial for $\zeta^{i}$ to be embedded in the hyperboloid when $d$ is greater than 2 where 
the advantage of embedding in Poincare space is reduced. We additionally observe improved stability and convergence of $\zeta^{i} \in \mathbb{H} ^ {M_i \times d,1}$.

\subsection{Hyperbolic reasoning module}
\label{subsec:reasoning}
The second component of our method, the hyperbolic reasoning module ($\mathcal{R}^l$), is similar to the HGCNN \cite{chami2019hyperbolic} but here we learn the edges of the graph (relations/conjunction between $\zeta^i$ to $\zeta^{i+1}$) using a binary function ($1 = edge, 0 = no\ edge$). This constructs a graph structure equivalent to a tree which is used for reasoning about Euclidean external representations of the visual world. 
Similar to \cite{chami2019hyperbolic}, the first stage of hyperbolic reasoning is a hyperbolic feature transformation shown in Equation \ref{eqn:ht}, which is performed in the unit hyperboloid where we found training to be more stable compared to within the Poincare unit disc/ball. 
Therefore we map a codebook $\zeta^i$ from Poincare space to the hyperboloid using Equation \ref{eqn:hp} before applying a hyperboloid linear layer ($h(x)^{\mathbb{H}, 1}$) to each codebook vector (Equation~\ref{eqn:ht}). This is followed by a logarithmic mapping (Equation~\ref{eqn:maph}) to $\mathcal{T}_o\mathbb{H}^{1,d}$. 
The second stage of hyperbolic reasoning is the aggregation of symbols in $\mathcal{T}_o\mathbb{H}^{1,d}$ as proposed in \cite{chami2019hyperbolic} to form $\zeta^{i+1}$. This is achieved by first learning the edges or equivalently the property of conjunction over symbols with a binary attention layer \cite{helwegen2019latent} in $\mathcal{R}^l$, with weights $w_{l}^{j,k} \in \mathbb{R}^{j \times k} \in \{0, 1\}, j = 0,1 \ldots M_i, k = 0,1 \ldots M_{i+1}$. 
We then perform mean aggregation for the merging of symbols with edges from $\zeta^i$ to $\zeta^{i+1}$ by matrix multiplying with weights matrix $a_{l} \in \mathbb{R}^{j \times k}$ before mapping back to Poincare space (Equation~\ref{eqn:pmt}). 
This aggregation is an approximation of the Freschet mean in hyperbolic space as assumed in \cite{chami2019hyperbolic}.
A single pass through $\mathcal{R}^l$ is summarised in the first line of Equation  \ref{eqn:hypoverall} below with the second %equation 
line calculating $a_{l}^{j,k}\in a_{l}$.
\begin{align}
    \begin{split}
     \zeta^{i+1}  &= exp_{o}^{\mathbb{B}, 1} ( a_{l}^\intercal \ log_{o}^{\mathbb{H}, 1} (h^{\mathbb{H}, 1}(\psi_{\mathbb{B}^{d, 1}\rightarrow\mathbb{H}^{d, 1}}(\zeta^{i}))))\\
     a_{l}^{j,k} &=
       \begin{cases}
         \frac{1}{\sum_{j=0}^{j = M_i} w_{l}^{j,k}}, & \text{if} \ w_{l}^{j,k} = 1 \\
         0, & \text{otherwise}
       \end{cases}
    \end{split}
    \label{eqn:hypoverall}
\end{align}

In order for the hyperbolic reasoning module to update its weights to form an accurate abstraction tree, it needs to classify the discretised samples $z_q$ from the Euclidean representation, $\mathcal{E}$. We firstly however apply a feature attention function in Euclidean space with continuous weights $w_l$ constrained between 0 and 1 to learn a reduced linear combination of the features in $z_{q}^{i}$. This is followed by a mapping to the Poincare unit ball formally defined as $z_{pq}^{i} = exp_{o}^{\mathbb{B},1} (w_{l}^\intercal z_{q}^{i})$ (see Equation~\ref{eqn:pmt}). Next, every $z_{pqk'}^{i} \in z_{pq}^{i}$ is mapped to the nearest codebook vector by Poincare distance in $\zeta^i$  using Equation \ref{eqn:pd} to form $z_{pq}^{i+1}$. Note, unlike in $\mathcal{VQ}$ for symbol formation, we do not directly update the symbols in $\zeta_i$ so that it is only learned by a function $R^l$ of $\zeta_{i-1}$ in order to faithfully learn an abstraction tree. Therefore, straight through gradient approximation is not required in this abstraction process.  We complete abstraction by moving our discrete sample back into Euclidean space to form $z_{q}^{i+1}$ by computing $log_{o}^{\mathbb{B},1}(z_{pq}^{i+1})$ (Equation~\ref{eqn:pmt}).
The feature attention function is necessary due to there being multiple edges from a single child symbol. Feature attention ensures the most likely symbol is sampled by being aware of the neighbouring symbols in $z_{q}^{i}$ which prevents the hierarchy from collapsing into one single discrete symbol. For example, in Figure \ref{fig:example}, $\zeta^{0}_{2}$ is related to $\zeta^{1}_{1}$ and $\zeta^{1}_{2}$; therefore assuming $z_{q}^{i}$ containing $\zeta^{0}_{2}$ also has $\zeta^{0}_{1}$ but not $\zeta^{0}_{3}$, then $\zeta^{0}_{2}$ maps to $\zeta^{1}_{1}$ as $\zeta^{0}_{1}$ is related to $\zeta^{1}_{1}$. 
% \akcom{TODO: need to write it's implications....}
% \begin{equation}
%   \label{eq:pcd}
%   \mathbb{P}(z_{pq}^{i+1} = k\mid z_{pq}^{i}) = \left\{
%     \begin{array}{ll}
%       1 \qquad   \text{if} \quad k = argmin_j (d^{\mathbb{B}, 1}(z_{pq}^{i}, \zeta^i))\\
%       0 \qquad   \text{otherwise}
%     \end{array}
%     \right\}
% \end{equation}

Finally, $z_{q}^{n}$ is mapped to the class prediction by average pooling to a linear class projection layer (Figure \ref{fig:overview}) to match the dimensionality of the output. 
The knowledge distillation loss is defined as the cross-entropy loss between the classifiers prediction ($y$) and the hyperbolic discrete surrogate model's prediction ($\hat{y}$): $\mathcal{L}_{dist}(\hat{y}, y)$.
We determine that the Poincare distances (Equation~\ref{eqn:pd}) between codebooks  correspond to graph distances in the tree. 
Therefore a Poincare codebook loss $\mathcal{L}_{Poincare}$ is calculated such that symbols with an edge are closer together and those without an edge are pushed apart in hyperbolic space. 
First, let $u$ to be any codebook vector in the set of all $\zeta^{i\ldots n}$ while $v$ and $v'$ are defined as codebook vectors with and without an edge with $u$ respectively, then two sets $\mathcal{P}$ and $\mathcal{W}$ are created such that; $u,v \in \mathcal{P}$ and $u,v' \in \mathcal{W}$. Given this, we can express the Poincare codebook loss as follows:  
% \akcom{what is D and N? it's not defined anywhere. $\mathcal{D}$ is used for dataset aswell, $\mathcal{N}$ is used for negative data samples in ILP reasoning....}
$\mathcal{L}_{Poincare} = \frac{\sum_{u,v \in \mathcal{P}} e^{d^{\mathbb{B}, 1}(u,v)}}{\sum_{u,v' \in \mathcal{W}} e^{d^{\mathbb{B}, 1}(u,v')}}$. We now define our total training loss as: $\mathcal{L}_{Total} = \mathcal{L}_{dist} + \mathcal{L}_{quant} + \mathcal{L}_{Poincare}$. 
% + \epsilon \mathcal{L}_{cb}, \ \epsilon \in \{0,1\}.

All continuous weights in our model framework are updated with Adam optimisation while the binary weights in $\mathcal{R}^l$ are updated using the \emph{Bop} algorithm proposed by \cite{helwegen2019latent}.  In this algorithm, the strength of gradient signal at time $t$ is determined by looking at the continuous exponential moving average $m_t$ of accumulated gradients up to the gradient, $g_t$ at $t$. The binary weights are then updated subjected to $m_t$ exceeding a threshold $\tau$ and the sign of $w_{t, l}^{j,k}$ matching $m_t$.
In our case, we map our weights to \{0, 1\} rather than \{-1, 1\}, by initializing weights randomly $\in \{0, 1\}$ and modify the update rule proposed in \cite{helwegen2019latent}.  This is shown in Equation \ref{eqn:binaryupdate} below, where the first %equation in \ref{eqn:binaryupdate} 
line corresponds to gradient strength where $\gamma$ is the adaptivity rate \cite{helwegen2019latent} and the second %equation 
line defines the update rule \cite{helwegen2019latent}. Please refer to appendix \ref{sec:appendix} for further training details.
% By following this update rule, we ensure that weights are either 0 or 1 while at the same time, following all the properties described in \cite{helwegen2019latent}. 
\begin{align}
    \begin{split}
       m_t &= (1 - \gamma)m_{t-1} + \gamma g_t \\
    %   \mathbb{E}_{e,y} [ \nabla_{w_t} (\mathcal{L}(\zeta(e), y))]\\
       w^{j,k}_{t,l} &= 
        \begin{cases}
            |w^{j,k}_{t-1,l} - 1|,    & \text{if } |m_t^i| > \tau \text{ and }\text{sign}(m_t^i) = \text{sign}(w_{t-1,l}^{j,k})\\
            w^{j,k}_{t-1,l},              & \text{otherwise}
        \end{cases}
    \end{split}
    \label{eqn:binaryupdate}
\end{align}

\subsection{Explanations and visual semantics}
\label{subsec:explanations}
\begin{wrapfigure}{r}{0.3\textwidth}
  \vspace{-20pt}
  \begin{center}
    \includegraphics[width=0.28\textwidth]{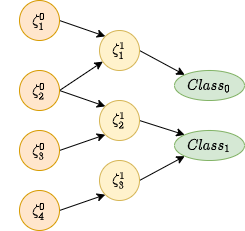}
    \hfill
    \caption{Example abstraction tree}
    \label{fig:example}
  \end{center}
  \vspace{-20pt}
  
\end{wrapfigure}

We now aim to generate a hierarchy of rules explaining the reasons behind the classifier's decision. 
At the end of training, all codebook symbols without an edge to or from itself are dropped and an abstraction tree is constructed based on the edges learnt between codebooks. 
% Once trained binary reasoning layer can be restructured as a tree, which describes the way of abstraction of symbols down the hierarchy.
We provide both global class-level and local example-level induced abstraction trees as a part of our explanation, along with the visual semantics for symbols in the example-level tree.

In Figure \ref{fig:example}, we have two codebooks $\zeta^0$ and $\zeta^1$ each with 4 and 3 symbols, which are getting abstracted to form the concept; $Class_0$ and $Class_1$.
% , demonstrates the above defined example.
% In the above example, it can be seen that symbols $\zeta^0_2$ and $\zeta^0_3$ are getting merged to form an abstract concept $\zeta^1_2$ which is further getting merged with $\zeta^1_3$ to form higher level abstraction of concept $Class_1$.
A class-level induced tree indicates the subtree corresponding to a specific class. 
For example, in \ref{fig:example}, the subtree for $Class_0$ consists of symbols, $\zeta_0^1, \zeta^0_2, \zeta^1_1$ and $Class_0$. An image level tree is input dependent, and forms a subtree of a class-level tree.

The extraction of visual rules, first requires us to train a decoder block to reconstruct images in Euclidean space as perceived by the classifier, with a reconstruction loss $\mathcal{L}_{recon} = ||\mathbb{D}(z_q) - x||_2^2$, where $z_q \sim \mathbb{C}$ and $x \sim \mathcal{X}$ described in figure \ref{fig:overview}.
During training, we make sure that the gradients from the decoder block do not affect the weights of the discrete surrogate model, to maintain faithfulness of the discretisation process.
% As described in Figure \ref{fig:overview}, the decoder block is trained only using features obtained from an Euclidean codebook  $\mathbb{E}$; this ensures the consistency of visual rules for every symbol across the hierarchy.
In order to explain the semantic meaning of any symbol; $\zeta_j^i \in z_q^i$, we must project the selected symbol onto $\mathbb{C}$ and visualise the semantic difference using a decoder $\mathbb{D}$.
In the first step to compute the semantic difference $\delta(\hat{x})$, we deduce all the symbols in $z_q^0 \sim \zeta^0$ responsible for activating $\zeta_j^i \in z_q^i$ using the image-level tree to form a set $\mathcal{S}  \subset z_q^0$.
$\zeta^0$ has a one to one mapping with $\mathbb{C}$ such that one can now form a new set $\mathcal{A} \subset  z_q$. 
We use $\mathcal{A}$ to perform a controlled intervention on $z_q \sim \mathbb{C}$ which is inputted into $\mathbb{D}$ to calculate $\delta(\hat{x})$. 
We can formally define this process as: $\delta(\hat{x}) = \mathbb{D}(z_q) - \mathbb{D}(z_q; do(z_{qk} = 0), \forall \quad z_{qk} \in \mathcal{A}$), where $z_{qk} \in z_q$.
% This allows to 
% As $\zeta^0$ has a one to one mapping with $\mathbb{E}$, 
% % As the decoder only uses features from $\zeta^0$ to explain any symbol in down stream codebooks, we first project the effect of any selected symbol onto $\zeta^0$ and visualise the effect. 
% % In practice the projection of a selected symbol is done by finding all the \emph{root-nodes} responsible for activating that node. 
% % For example in figure \ref{fig:example} projection of $\zeta^1_2$ correspond to $\{ \zeta^0_2, \zeta^0_3 \}$. 
% to compute the semantic difference $\delta(\hat{x})$ for a symbol $\zeta_j^i $, we first find all the symbols for $z_q^0 \sim \zeta^0$ responsible for activating $\zeta_j^i$ and perform controlled intervention on $\mathbb{E}$ formally defined as: $\delta(\hat{x}) = \mathbb{D}(z_q) - \mathbb{D}(z_q; do(\mathbb{E}_{\mathcal{S}} = 0))$.
% For example, in Figure \ref{fig:example}, the semantic meaning of a symbol $\zeta_2^1$ is same as the collective semantic meaning of symbols $\{ \zeta^0_2, \zeta^0_3 \}$.
% %which also corresponds to $\{\mathbb{E}_2, \mathbb{E}_3 \}$.

% defined in equation\ref{eqn:effect}
% \begin{equation}
%     \mathcal{Y}(x) = \mathbb{D}(\zeta^0(f(x))) - \mathbb{D}(\zeta^0(f(x)); do(\zeta^0_i = 0))
%     \label{eqn:effect}
% \end{equation}
% \begin{equation}
%     \delta(\hat{x}) = \mathbb{D}(z_q) - \mathbb{D}(z_q); do(\mathbb{E}_i = 0)
%     \label{eqn:effect}
% \end{equation}

\section{Experiments}
\begin{wrapfigure}{l}{0.27\textwidth}
  \vspace{-20pt}

  \begin{center}
  
    \includegraphics[width=0.28\textwidth]{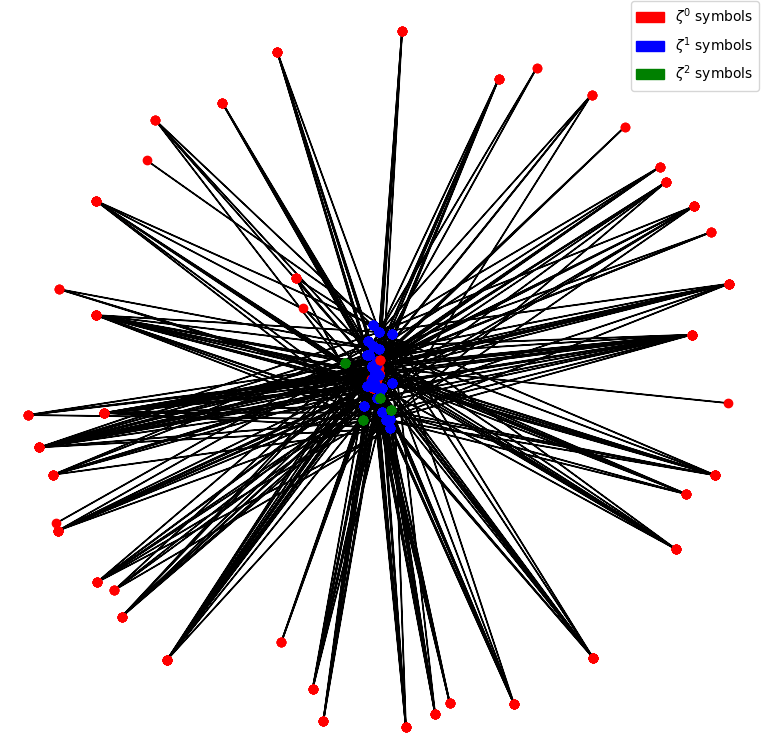}
    % \hspace
    \caption{2D Poincare embedding of symbols obtained for MNIST. Red, blue, and green nodes indicate symbols from $\zeta^0, \zeta^1, \zeta^2$ respectively}
    \label{fig:poincare}
  \end{center}
  \vspace{-10pt}

\end{wrapfigure}

We use our framework to explain two pre-trained classifiers, one trained on the MNIST dataset \cite{deng2012mnist} as a proof of concept study and the other trained on the AFHQ dataset \cite{choi2020stargan} to indicate the scalability and generalisability of our proposed method. In the following experiments, we choose a 3 level hierarchy. We determine through our ablations, the number of codebook vectors for each level $M_i$, to be 128, 32, and 4 for the MNIST experiment and 256, 64, and 2 for AFHQ. Please refer to appendix \ref{sec:appendix} for information on classifier models and codebook ablations.

We hypothesise that hyperbolic embeddings will better embed a hierarchical tree without distortion and hence allow to reduce the dimensionality $d$ of $\zeta^i$ such that knowledge distillation would not be affected. 
We support this hypothesis by achieving  better knowledge distillation accuracy with Poincare embeddings highlighted in table \ref{table:hyp}.  One also notes increasingly better performance as we reduce the dimensionality of $\zeta^i$ down to 2 on the MNIST and AFHQ dataset. 
Figure \ref{fig:poincare} shows the 2 dimensional embeddings on the Poincare disk for the MNIST dataset maintaining a robust hierarchy. 
% \akcom{needs some clarification}

\begin{wrapfigure}[22]{r}{0.32\textwidth}
  \vspace{-20pt}
  \begin{center}
    \includegraphics[width=0.31\textwidth]{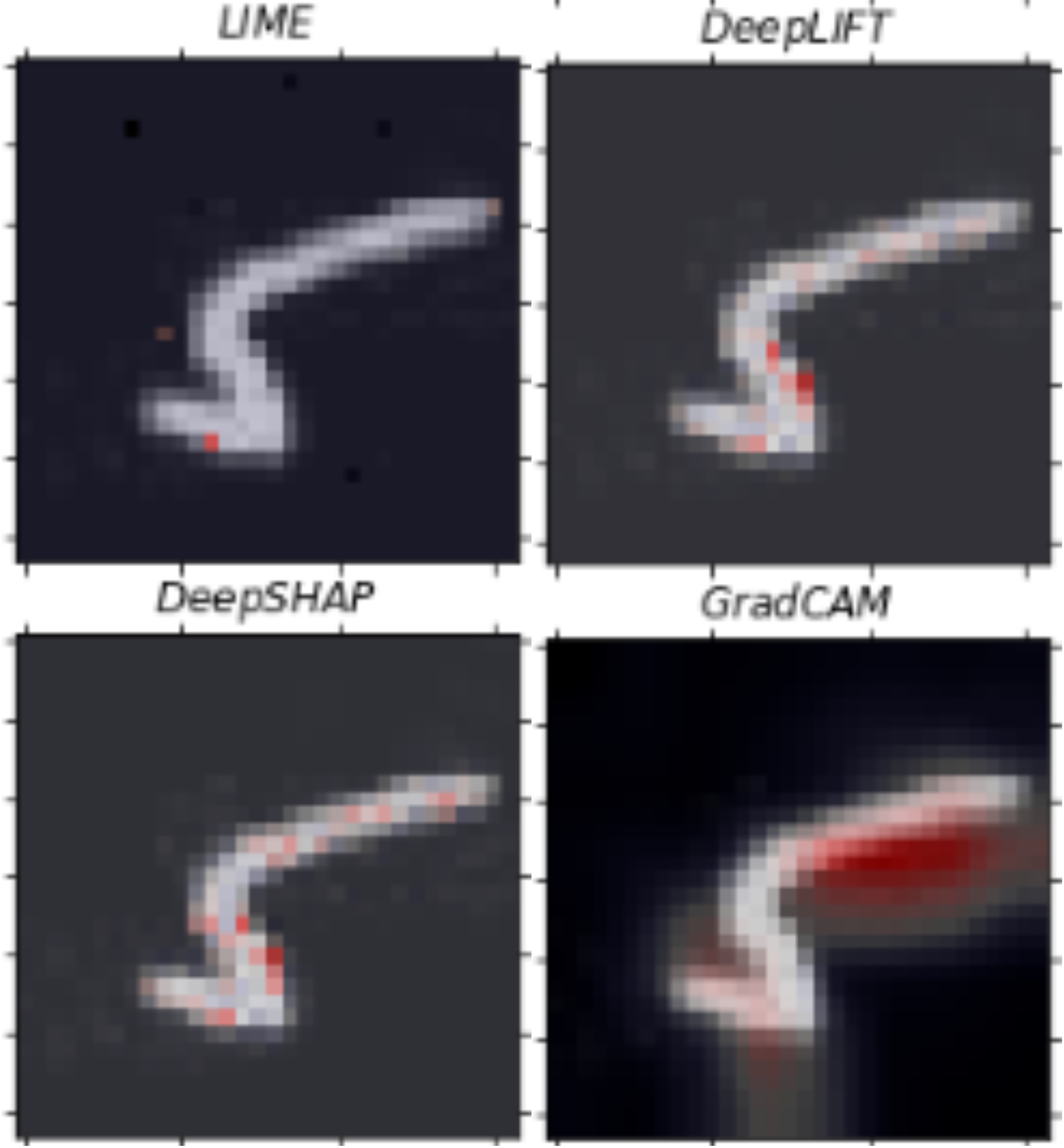}
    % \hspace
    % \vspace{10pt}
    \caption{Illustration of existing explanation techniques, top-left, top-right, bottom-left, and bottom-right corresponds to LIME, DeepLIFT, DeepSHAP, and GradCAM explanations respectively.}
    \label{fig:comptest}
  \end{center}
  \vspace{-10pt}
\end{wrapfigure}

% We selected $32 \times 32$ MNIST due to its simplicity; it makes it easy to establish our points and easily compare other existing methods, while $128 \times 128$ AFHQ is a high resolution animal faces dataset; hierarchical explanations on AFHQ .
Next, we qualitatively compare the explanations obtained from our framework against standard explainability frameworks: LIME \cite{lime}, DeepSHAP \cite{SHAP}, deepLIFT \cite{deeplift}, and gradCAM \cite{selvaraju2017grad}. Figure \ref{fig:comptest} describes the explanations obtained from standard existing methods. It is clear that most of the existing explanation methods focus on either pixel importance or gradient-based attention and do not differentiate the effect of multiple concepts. These explanations also are quite limited without yielding any sort of reasoning.

% \begin{figure}
%   \centering
%     \subfloat[]{\includegraphics[width=.38\textwidth]{imgs/comptest.png}} \hfill
%     \subfloat[]{\includegraphics[width=.38\textwidth]{imgs/poincaresphere-0-mnistpoincare.png}} \hfill
%     \caption{caption}
%     \label{fig:comptest}
% \end{figure}

% Attention based explanation methods also cannot differentiate the effect of multiple concepts involved in making certain predictions. 
% As these explanations are obtained in euclidean space, the meaning of semantic distance is ill-defined, i.e., these methods fail to differentiate between cat ears to dog ears since the semantic distance of these two different concepts cannot be gauged effectively in euclidean space.

% As previously described attention based methods cannot differentiate between concepts inside the region of interest (like cat-ears and dog-ears).

\begin{figure}[hbt!]
    \vspace{-12pt}
    \centering
    \subfloat[Class-level tree]{\includegraphics[width=.29\textwidth]{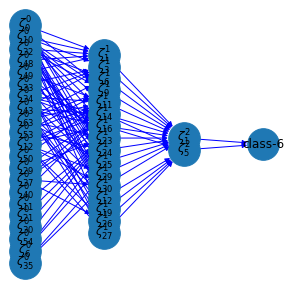}} \hfill
    \subfloat[Image-level tree]{\includegraphics[width=.29\textwidth]{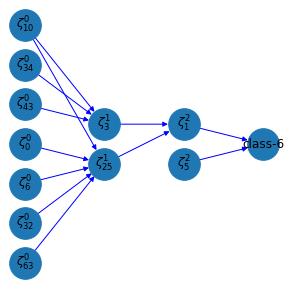}} \hfill
    \subfloat[Visual rules]{\includegraphics[width=.40\textwidth]{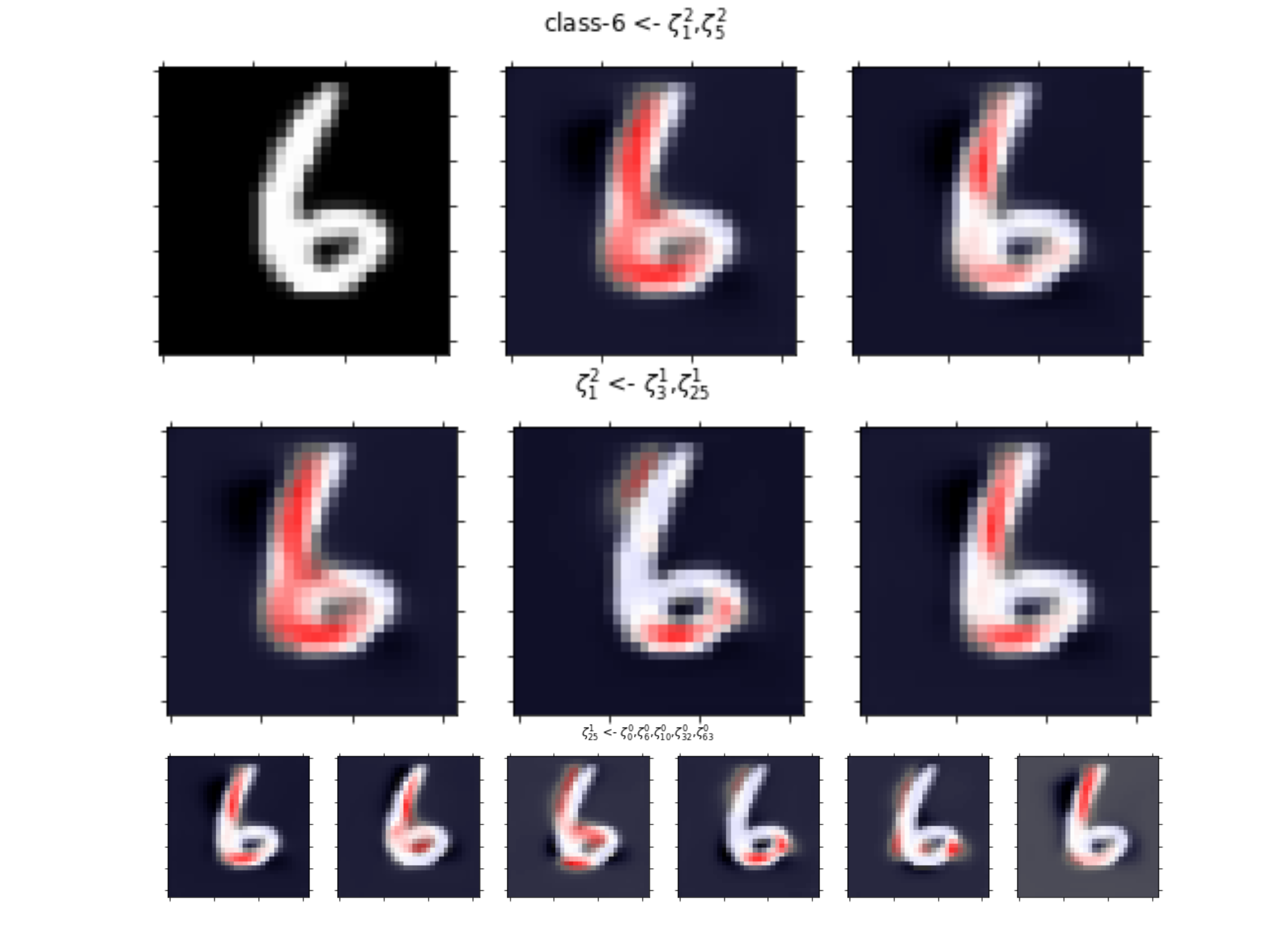}} \hfill
    % \vspace{10pt}
    \caption{This figure describes the explanations obtained using the proposed framework for a MNIST classifier. $\zeta^i \in \mathbb{B}^{M_i \times 16,1}$(a) Demonstrates the obtained class-level tree, which is a complete set of symbols responsible for entailing class; '6', (b) indicates a image-level tree with symbols responsible for making a decision for a given image, and (c) demonstrates the some of the visual rules obtained from a image-level tree.}
    \label{fig:mnistexp}
     \vspace{-12pt}
\end{figure}

Our proposed method go beyond feature attribution by allowing the user to decide on the level of abstractness upon which to provide the symbolic and corresponding visual semantic rules which make up explanations.
Figure \ref{fig:mnistexp} demonstrates the effectiveness of our explanations on a pre-trained classifier for MNIST.
Figure \ref{fig:mnistexp}(a) shows the class-level \emph{global-tree} representing our trained reasoning blocks; this tree includes all possible symbol abstractions for a particular class. To explain a given image we construct a image-level \emph{local-tree} which is described in Figure \ref{fig:mnistexp}(b). 
Figure \ref{fig:mnistexp}(c) provides a visual description of an image-level tree where we plot a single visual rule from each abstract level of reasoning.
In this example, the visual semantics for a rule obtained for a given image $\mathrm{x}$ corresponding to $\mathrm{Class6(x)} \leftarrow \zeta_5^2(\mathrm{x}), \zeta_1^2(\mathrm{x})$ is described in the first row of Figure \ref{fig:mnistexp}(c). The second row of Figure \ref{fig:mnistexp}(c) visualises the obtained rule corresponding to $\zeta_1^2(\mathrm{x}) \leftarrow \zeta_3^1(\mathrm{x}), \zeta_{25}^1(\mathrm{x})$ while the last row visualises how $\zeta^1_{25}$ is obtained using symbols from $\zeta^0$. As we go down the level of abstraction or in other terms, as we move closer to the boundary of Poincare space \ref{fig:poincare}, visually the symbols start to move from a complete digit heatmap to a more focused region in a digit, demonstrating a visual hierarchical explanation. One can also note the distinct semantic differences between individual symbols highlighting the advantage of discretising features.
Similar explanations and visual semantic behaviour can be observed for a model trained on the AFHQ dataset shown in Figure \ref{fig:afhqexp}  where we again observe symbols getting localized as we go deeper into the hierarchy. Please refer to appendix \ref{sec:appendix} for further details and examples.

\begin{table}[hbt!]
\vspace{-10pt}
\centering
\caption{Knowledge distillation accuracy of different dimensional Euclidean and Poincare embeddings on the MNIST and AFHQ classifiers.}
\vspace{10pt}
\label{table:hyp}
\begin{tabular}{@{}ccccccc@{}}
\toprule
\multirow{2}{*}{\textbf{\begin{tabular}[c]{@{}c@{}}Embedding dim $\rightarrow$ \ \\ Dataset $\downarrow$\end{tabular}}} & \multicolumn{3}{c}{\textbf{Poincare}}         & \multicolumn{3}{c}{\textbf{Euclidean}}   \\ \cmidrule(l){2-7} 
                                                                                                                                    & \textbf{2}    & \textbf{4}    & \textbf{16}   & \textbf{2}    & \textbf{4} & \textbf{16} \\ \cmidrule(l){1-7}
MNIST                                                                                                                               & \textbf{0.90} & \textbf{0.96} & \textbf{0.99} & 0.81 & 0.92       & 0.95        \\
AFHQ                                                                                                                                & \textbf{0.90} & \textbf{0.95} & \textbf{0.98} & 0.80 & 0.90       & 0.97        \\ \bottomrule
\end{tabular}
\end{table}

% indicates how symbol $\zeta_1^2$ was generated using previous codebook, in this case the

\begin{figure}[hbt!]
     \vspace{-12pt}
    \centering
    \subfloat[Class-level tree]{\includegraphics[width=.29\textwidth]{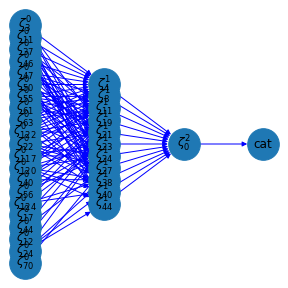}} \hfill
    \subfloat[Image-level tree]{\includegraphics[width=.29\textwidth]{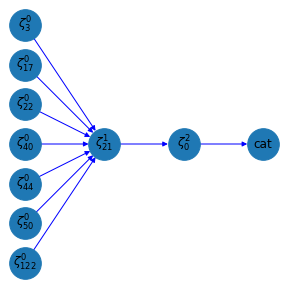}} \hfill
    \subfloat[Visual rules]{\includegraphics[width=.30\textwidth]{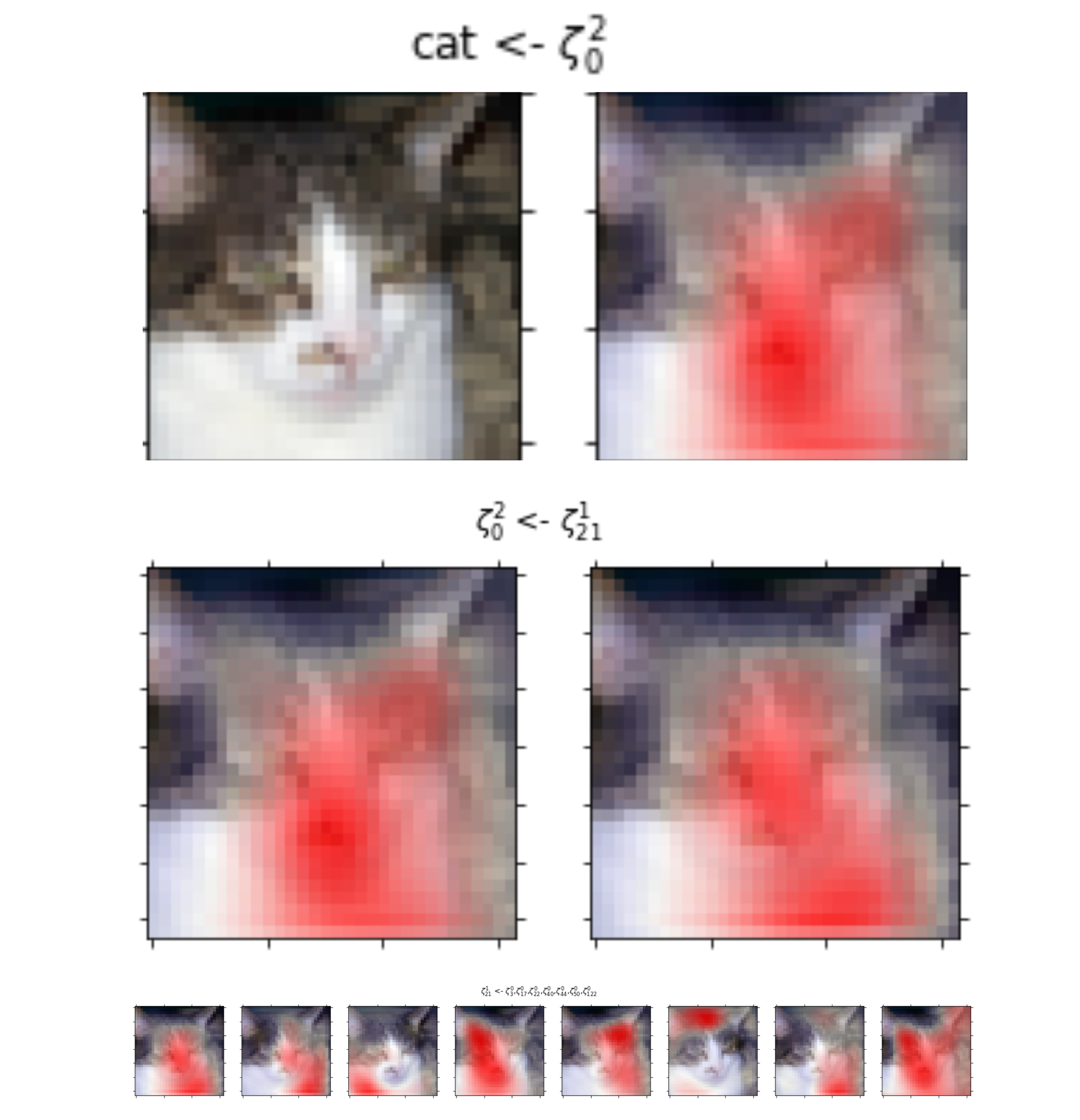}} \hfill
    % \vspace{10pt}
    \caption{This figure describes the explanations obtained using the proposed framework for a AFHQ classifier for the class; 'cat'. $\zeta^i \in \mathbb{B}^{M_i \times 16,1}$}
    \label{fig:afhqexp}
    \vspace{-12pt}
\end{figure}

% this tree indicates how the symbols are getting merged to create symbols in downstream codebooks.
% As figure \ref{fig:mnistexp}(a) is a complete tree

% \ref{sec:appendix}.

\section{Conclusion}
This work provides novel hierarchical explanations for deep discriminative models, which was demonstrated on the MNIST and AFHQ datasets.
The proposed framework discretises the continuous latent space of classifiers into $M_0$ discrete features, followed by multiple layers of reasoning and discrete abstraction in hyperbolic space to form an abstraction tree which provide hierarchical rules as explanations for the Euclidean visual world. We demonstrate hyperbolic geometry allows to embed our knowledge tree with minimal distortion compared to the Euclidean counterpart.
% We perform all the operations of sampling and abstraction in hyperbolic space to form an internal representation of Euclidean visual world which preserve semantic distances and hierarchy between symbolic features to in an image, which results in learning symbols that are hierarchical and responsible for specific class confidence.
The results show the existence of consistent and distinct symbolic hierarchical rules for each class which is visualised by generating attention regions in an image.
% % In this work, we did not explore the possibility of assigning semantic meaning to leaned symbols.
% We believe this method can be used to provide high-level explanations for the model as well as . 
For future work, our framework can be developed into a stand-alone deep discriminative neuro-symbolic model which improves generalisability as well as interpretability. We can here also explore the advantage of learnable curvature similar to \cite{chami2019hyperbolic} to embed symbolic data in hyperbolic space.
We plan to extend this method with domain experts to assign semantic meaning to symbols and extract high-level descriptive explanations as well as highlight biases and inconsistencies in features learned by a given model.

\noindent \textbf{Acknowledgements.} This work was supported and funded by Cancer Research UK (CRUK) (C309/A28804) and UKRI centre for Doctoral Training in Safe and Trusted AI (EP/S023356/1).

\bibliographystyle{unsrt}
\bibliography{main.bib}

\begin{thebibliography}{10}

\bibitem{doshi2017role}
Finale Doshi-Velez, Ryan Budish, and Mason Kortz.
\newblock The role of explanation in algorithmic trust.
\newblock Technical report, Technical report, Artificial Intelligence and
  Interpretability Working Group~…, 2017.

\bibitem{kroll2015accountable}
Joshua~Alexander Kroll.
\newblock {\em Accountable algorithms}.
\newblock PhD thesis, Princeton University, 2015.

\bibitem{kim2018interpretability}
Been Kim, Martin Wattenberg, Justin Gilmer, Carrie Cai, James Wexler, Fernanda
  Viegas, et~al.
\newblock Interpretability beyond feature attribution: Quantitative testing
  with concept activation vectors (tcav).
\newblock In {\em International conference on machine learning}, pages
  2668--2677. PMLR, 2018.

\bibitem{lipton2018mythos}
Zachary~C Lipton.
\newblock The mythos of model interpretability: In machine learning, the
  concept of interpretability is both important and slippery.
\newblock {\em Queue}, 16(3):31--57, 2018.

\bibitem{rightforright}
Wolfgang Stammer, Patrick Schramowski, and Kristian Kersting.
\newblock Right for the right concept: Revising neuro-symbolic concepts by
  interacting with their explanations.
\newblock In {\em Proceedings of the IEEE/CVF Conference on Computer Vision and
  Pattern Recognition}, pages 3619--3629, 2021.

\bibitem{tellmewhy}
Andrew~K Lampinen, Nicholas~A Roy, Ishita Dasgupta, Stephanie~CY Chan,
  Allison~C Tam, James~L McClelland, Chen Yan, Adam Santoro, Neil~C Rabinowitz,
  Jane~X Wang, et~al.
\newblock Tell me why!--explanations support learning of relational and causal
  structure.
\newblock {\em arXiv preprint arXiv:2112.03753}, 2021.

\bibitem{debate}
Geoffrey Irving, Paul Christiano, and Dario Amodei.
\newblock Ai safety via debate.
\newblock {\em arXiv preprint arXiv:1805.00899}, 2018.

\bibitem{lime}
Marco~Tulio Ribeiro, Sameer Singh, and Carlos Guestrin.
\newblock " why should i trust you?" explaining the predictions of any
  classifier.
\newblock In {\em Proceedings of the 22nd ACM SIGKDD international conference
  on knowledge discovery and data mining}, pages 1135--1144, 2016.

\bibitem{SHAP}
Scott~M Lundberg and Su-In Lee.
\newblock A unified approach to interpreting model predictions.
\newblock In I.~Guyon, U.~V. Luxburg, S.~Bengio, H.~Wallach, R.~Fergus,
  S.~Vishwanathan, and R.~Garnett, editors, {\em Advances in Neural Information
  Processing Systems}, volume~30. Curran Associates, Inc., 2017.

\bibitem{deeplift}
Avanti Shrikumar, Peyton Greenside, and Anshul Kundaje.
\newblock Learning important features through propagating activation
  differences.
\newblock In {\em International conference on machine learning}, pages
  3145--3153. PMLR, 2017.

\bibitem{selvaraju2017grad}
Ramprasaath~R Selvaraju, Michael Cogswell, Abhishek Das, Ramakrishna Vedantam,
  Devi Parikh, and Dhruv Batra.
\newblock Grad-cam: Visual explanations from deep networks via gradient-based
  localization.
\newblock In {\em Proceedings of the IEEE international conference on computer
  vision}, pages 618--626, 2017.

\bibitem{gradcampp}
Aditya Chattopadhay, Anirban Sarkar, Prantik Howlader, and Vineeth~N
  Balasubramanian.
\newblock Grad-cam++: Generalized gradient-based visual explanations for deep
  convolutional networks.
\newblock In {\em 2018 IEEE winter conference on applications of computer
  vision (WACV)}, pages 839--847. IEEE, 2018.

\bibitem{integratedcam}
Sam Sattarzadeh, Mahesh Sudhakar, Konstantinos~N Plataniotis, Jongseong Jang,
  Yeonjeong Jeong, and Hyunwoo Kim.
\newblock Integrated grad-cam: Sensitivity-aware visual explanation of deep
  convolutional networks via integrated gradient-based scoring.
\newblock In {\em ICASSP 2021-2021 IEEE International Conference on Acoustics,
  Speech and Signal Processing (ICASSP)}, pages 1775--1779. IEEE, 2021.

\bibitem{goyal2019counterfactual}
Yash Goyal, Ziyan Wu, Jan Ernst, Dhruv Batra, Devi Parikh, and Stefan Lee.
\newblock Counterfactual visual explanations.
\newblock In {\em International Conference on Machine Learning}, pages
  2376--2384. PMLR, 2019.

\bibitem{tace}
Amirata Ghorbani, James Wexler, James Zou, and Been Kim.
\newblock Towards automatic concept-based explanations.
\newblock {\em arXiv preprint arXiv:1902.03129}, 2019.

\bibitem{dissect}
Asma Ghandeharioun, Been Kim, Chun-Liang Li, Brendan Jou, Brian Eoff, and
  Rosalind~W Picard.
\newblock Dissect: Disentangled simultaneous explanations via concept
  traversals.
\newblock {\em arXiv preprint arXiv:2105.15164}, 2021.

\bibitem{concepttrail}
Avinash Kori, Parth Natekar, Balaji Srinivasan, and Ganapathy Krishnamurthi.
\newblock Interpreting deep neural networks for medical imaging using concept
  graphs.
\newblock In {\em International Workshop on Health Intelligence}, pages
  201--216. Springer, 2021.

\bibitem{sauer2021counterfactual}
Axel Sauer and Andreas Geiger.
\newblock Counterfactual generative networks.
\newblock {\em arXiv preprint arXiv:2101.06046}, 2021.

\bibitem{chang2018explaining}
Chun-Hao Chang, Elliot Creager, Anna Goldenberg, and David Duvenaud.
\newblock Explaining image classifiers by counterfactual generation.
\newblock {\em arXiv preprint arXiv:1807.08024}, 2018.

\bibitem{nemirovsky2020countergan}
Daniel Nemirovsky, Nicolas Thiebaut, Ye~Xu, and Abhishek Gupta.
\newblock Countergan: Generating realistic counterfactuals with residual
  generative adversarial nets.
\newblock {\em arXiv preprint arXiv:2009.05199}, 2020.

\bibitem{lang2021explaining}
Oran Lang, Yossi Gandelsman, Michal Yarom, Yoav Wald, Gal Elidan, Avinatan
  Hassidim, William~T Freeman, Phillip Isola, Amir Globerson, Michal Irani,
  et~al.
\newblock Explaining in style: Training a gan to explain a classifier in
  stylespace.
\newblock {\em arXiv preprint arXiv:2104.13369}, 2021.

\bibitem{armstrong1983some}
Sharon~Lee Armstrong, Lila~R Gleitman, and Henry Gleitman.
\newblock What some concepts might not be.
\newblock {\em Cognition}, 13(3):263--308, 1983.

\bibitem{burnston2021evolving}
Daniel~C Burnston and Philipp Haueis.
\newblock Evolving concepts of “hierarchy” in systems neuroscience.
\newblock In {\em Neural Mechanisms}, pages 113--141. Springer, 2021.

\bibitem{hudson2019learning}
Drew Hudson and Christopher~D Manning.
\newblock Learning by abstraction: The neural state machine.
\newblock {\em Advances in Neural Information Processing Systems}, 32, 2019.

\bibitem{vedantam2019probabilistic}
Ramakrishna Vedantam, Karan Desai, Stefan Lee, Marcus Rohrbach, Dhruv Batra,
  and Devi Parikh.
\newblock Probabilistic neural symbolic models for interpretable visual
  question answering.
\newblock In {\em International Conference on Machine Learning}, pages
  6428--6437. PMLR, 2019.

\bibitem{mao2019neuro}
Jiayuan Mao, Chuang Gan, Pushmeet Kohli, Joshua~B Tenenbaum, and Jiajun Wu.
\newblock The neuro-symbolic concept learner: Interpreting scenes, words, and
  sentences from natural supervision.
\newblock {\em arXiv preprint arXiv:1904.12584}, 2019.

\bibitem{yi2018neural}
Kexin Yi, Jiajun Wu, Chuang Gan, Antonio Torralba, Pushmeet Kohli, and Josh
  Tenenbaum.
\newblock Neural-symbolic vqa: Disentangling reasoning from vision and language
  understanding.
\newblock {\em Advances in neural information processing systems}, 31, 2018.

\bibitem{tee2020information}
James Tee and Desmond~P Taylor.
\newblock Is information in the brain represented in continuous or discrete
  form?
\newblock {\em IEEE Transactions on Molecular, Biological and Multi-Scale
  Communications}, 6(3):199--209, 2020.

\bibitem{vetter2014varieties}
Petra Vetter and Albert Newen.
\newblock Varieties of cognitive penetration in visual perception.
\newblock {\em Consciousness and cognition}, 27:62--75, 2014.

\bibitem{wessinger2001hierarchical}
CM~Wessinger, J~VanMeter, Biao Tian, J~Van~Lare, James Pekar, and Josef~P
  Rauschecker.
\newblock Hierarchical organization of the human auditory cortex revealed by
  functional magnetic resonance imaging.
\newblock {\em Journal of cognitive neuroscience}, 13(1):1--7, 2001.

\bibitem{meunier2009hierarchical}
David Meunier, Renaud Lambiotte, Alex Fornito, Karen Ersche, and Edward~T
  Bullmore.
\newblock Hierarchical modularity in human brain functional networks.
\newblock {\em Frontiers in neuroinformatics}, 3:37, 2009.

\bibitem{glanois2021neuro}
Claire Glanois, Xuening Feng, Zhaohui Jiang, Paul Weng, Matthieu Zimmer, Dong
  Li, and Wulong Liu.
\newblock Neuro-symbolic hierarchical rule induction.
\newblock {\em arXiv preprint arXiv:2112.13418}, 2021.

\bibitem{mitchener2022detect}
Ludovico Mitchener, David Tuckey, Matthew Crosby, and Alessandra Russo.
\newblock Detect, understand, act: A neuro-symbolic hierarchical reinforcement
  learning framework.
\newblock {\em Machine Learning}, pages 1--27, 2022.

\bibitem{sharma2005probabilistic}
Rita Sharma and David Poole.
\newblock Probabilistic reasoning with hierarchically structured variables.
\newblock In {\em IJCAI}, pages 1391--1397. Citeseer, 2005.

\bibitem{van2017neural}
Aaron Van Den~Oord, Oriol Vinyals, et~al.
\newblock Neural discrete representation learning.
\newblock {\em Advances in neural information processing systems}, 30, 2017.

\bibitem{nickel2014reducing}
Maximilian Nickel, Xueyan Jiang, and Volker Tresp.
\newblock Reducing the rank in relational factorization models by including
  observable patterns.
\newblock {\em Advances in Neural Information Processing Systems}, 27, 2014.

\bibitem{sala2018representation}
Frederic Sala, Chris De~Sa, Albert Gu, and Christopher R{\'e}.
\newblock Representation tradeoffs for hyperbolic embeddings.
\newblock In {\em International conference on machine learning}, pages
  4460--4469. PMLR, 2018.

\bibitem{nickel2017poincare}
Maximillian Nickel and Douwe Kiela.
\newblock Poincar{\'e} embeddings for learning hierarchical representations.
\newblock {\em Advances in neural information processing systems}, 30, 2017.

\bibitem{ganea2018hyperbolic}
Octavian Ganea, Gary B{\'e}cigneul, and Thomas Hofmann.
\newblock Hyperbolic neural networks.
\newblock {\em Advances in neural information processing systems}, 31, 2018.

\bibitem{chami2019hyperbolic}
Ines Chami, Zhitao Ying, Christopher R{\'e}, and Jure Leskovec.
\newblock Hyperbolic graph convolutional neural networks.
\newblock {\em Advances in neural information processing systems}, 32, 2019.

\bibitem{liu2019hyperbolic}
Qi~Liu, Maximilian Nickel, and Douwe Kiela.
\newblock Hyperbolic graph neural networks.
\newblock {\em Advances in Neural Information Processing Systems}, 32, 2019.

\bibitem{muggleton2018meta}
Stephen Muggleton, Wang-Zhou Dai, Claude Sammut, Alireza Tamaddoni-Nezhad, Jing
  Wen, and Zhi-Hua Zhou.
\newblock Meta-interpretive learning from noisy images.
\newblock {\em Machine Learning}, 107(7):1097--1118, 2018.

\bibitem{lin2014bias}
Dianhuan Lin, Eyal Dechter, Kevin Ellis, Joshua~B Tenenbaum, and Stephen~H
  Muggleton.
\newblock Bias reformulation for one-shot function induction.
\newblock 2014.

\bibitem{michie1988machine}
Donald Michie.
\newblock Machine learning in the next five years.
\newblock In {\em Proceedings of the 3rd European conference on European
  working session on learning}, pages 107--122, 1988.

\bibitem{muggleton1995inverse}
Stephen Muggleton.
\newblock Inverse entailment and progol.
\newblock {\em New generation computing}, 13(3):245--286, 1995.

\bibitem{metagol}
Andrew Cropper and Stephen~H. Muggleton.
\newblock Metagol system.
\newblock https://github.com/metagol/metagol, 2016.

\bibitem{evans2018learning}
Richard Evans and Edward Grefenstette.
\newblock Learning explanatory rules from noisy data.
\newblock {\em Journal of Artificial Intelligence Research}, 61:1--64, 2018.

\bibitem{blockeel1998top}
Hendrik Blockeel and Luc De~Raedt.
\newblock Top-down induction of first-order logical decision trees.
\newblock {\em Artificial intelligence}, 101(1-2):285--297, 1998.

\bibitem{helwegen2019latent}
Koen Helwegen, James Widdicombe, Lukas Geiger, Zechun Liu, Kwang-Ting Cheng,
  and Roeland Nusselder.
\newblock Latent weights do not exist: Rethinking binarized neural network
  optimization.
\newblock {\em Advances in neural information processing systems}, 32, 2019.

\bibitem{deng2012mnist}
Li~Deng.
\newblock The mnist database of handwritten digit images for machine learning
  research [best of the web].
\newblock {\em IEEE signal processing magazine}, 29(6):141--142, 2012.

\bibitem{choi2020stargan}
Yunjey Choi, Youngjung Uh, Jaejun Yoo, and Jung-Woo Ha.
\newblock Stargan v2: Diverse image synthesis for multiple domains.
\newblock In {\em Proceedings of the IEEE/CVF conference on computer vision and
  pattern recognition}, pages 8188--8197, 2020.

\bibitem{zhang2016riemannian}
Hongyi Zhang, Sashank J~Reddi, and Suvrit Sra.
\newblock Riemannian svrg: Fast stochastic optimization on riemannian
  manifolds.
\newblock {\em Advances in Neural Information Processing Systems}, 29, 2016.

\bibitem{he2016identity}
Kaiming He, Xiangyu Zhang, Shaoqing Ren, and Jian Sun.
\newblock Identity mappings in deep residual networks.
\newblock In {\em European conference on computer vision}, pages 630--645.
  Springer, 2016.

\bibitem{iandola2014densenet}
Forrest Iandola, Matt Moskewicz, Sergey Karayev, Ross Girshick, Trevor Darrell,
  and Kurt Keutzer.
\newblock Densenet: Implementing efficient convnet descriptor pyramids.
\newblock {\em arXiv preprint arXiv:1404.1869}, 2014.

\end{thebibliography}
\newpage
\section{Appendix}
\label{sec:appendix}

\subsection{Hyperbolic mappings}

We show here the general equations $exp_{x}^{K}(y)$ and $log_{x}^{K}(v)$ with any negative curvature, $-1/K$ for $ \mathbb{H}^{d, K}$ in Equation \ref{eqn:ghe} and \ref{eqn:ghl} respectively as well as for $ \mathbb{B}^{d, K}$  in Equation \ref{eqn:gpe} and \ref{eqn:gpl} respectively \cite{chami2019hyperbolic, ganea2018hyperbolic}.

\subsubsection{Hyperboloid}
\begin{equation}
    exp_{x}^{\mathbb{H}, K}(y) = cosh\bigg(\frac{\lVert y \rVert _S}{\sqrt{K}}\bigg)\ x +  \sqrt{K}sinh\bigg(\frac{\lVert y \rVert _S}{\sqrt{K}}\bigg)\frac{y}{\lVert y \rVert _S} 
    \label{eqn:ghe}
\end{equation}

\begin{equation}
    log_{x}^{\mathbb{H}, K}(v) =  d^{\mathbb{H}, K} (x,v) \bigg(\frac{v+\frac{1}{K}\langle x, v \rangle _S \ x}{\lvert \lvert v+ \frac{1}{K} \langle x, v \rangle _S \ x\rvert \rvert_S}\bigg)
    \label{eqn:ghl}
\end{equation}

\subsubsection{Poincare}

A Riemmanian metric tensor is conformal to another Riemannian metric if it defines the same angles. In the case of the Poincare unit ball, there is a smooth conformal mapping, $\lambda:\mathbb{B} \rightarrow \mathbb{R}$ between the Euclidean metric tensor $g_{x}^{\mathbb{R}}$ and Poincare metric tensor $g_{x}^{\mathbb{B},1}$. The Poincare conformal factor $\lambda_{x} = \frac{2}{1 - \lVert x \rVert ^2}$ is defined such that $g_{x}^{\mathbb{B},1} = \lambda_{x}^{2}g_{x}^{\mathbb{R}}$.
\begin{equation}
    exp_{x}^{\mathbb{B},K}(y) = x \oplus^K \bigg(tanh(\frac{\sqrt{K} \lambda_x^K \lVert y \rVert}{2})\frac{y}{\sqrt{K}\lVert y \rVert }\bigg)
    \label{eqn:gpe}
\end{equation}

\begin{equation}
     log_{x}^{\mathbb{B},K}(v) =  \frac{2}{\sqrt{K} \lambda_x^K} arctanh(\sqrt{K}\lVert -x \oplus^K  v \rVert) \frac{-x \oplus^K v }{\lVert -x \oplus^K  v \rVert})
     \label{eqn:gpl}
\end{equation}

\subsection{Parallel transport}
In this section, we reveal the equations of parallel transport for vector $b$ on the tangential space of the origin to the tangential space of $v$. Formally, we show $P_{o \rightarrow v}^K(b)$ in $\mathbb{B}^{K,d}$ and $\mathbb{H}^{K,d}$ below.

\subsubsection{Hyperboloid}
The general form of this equation, $P_{x \rightarrow v}^K(b)$ shown below in Equation \ref{eqn:pth1} is derived in \cite{chami2019hyperbolic}.
\begin{equation}
    P_{o \rightarrow v}^K(b)  = b - \frac{\langle log_o(v), b \rangle _S}{d^{\mathbb{H}, 1}(o,v)^2}(log_o(v) + log_v(o)) 
    % \qquad \qquad \qquad \qquad \
    \label{eqn:pth1}
\end{equation}  
% \begin{equation}
%     P_{o \rightarrow v}^K(b)  = b - \bigg(-\lvert \lvert v_{1:d} \rvert \rvert _2, \frac{v_{1:d}}{\lvert \lvert v_{1:d} \rvert \rvert _2} \sqrt{K} - v_{0}\bigg)\bigg(\sum _{j = 1}^{j = d}  \frac{v_{1:d}}{\sqrt{K} \lvert \lvert v_{1:d} \rvert \rvert _2} b_{1:d} \bigg)
%     \label{eqn:pth2}
% \end{equation}

\subsubsection{Poincare}
Equation \ref{eqn:ptp} is derived in \cite{ganea2018hyperbolic}.
\begin{equation}
    P_{o \rightarrow v}^K(b)  = \frac{\lambda_o^K}{\lambda_v^K}b
    \label{eqn:ptp}
\end{equation}

\subsection{Projections}
We need to apply projections to constrain points to the manifolds and its tangential space after optimising. In our work, all operations are performed in the tangential plane and therefore one needs to only apply projections after the exponential mapping from the tangential space to the manifold which we define below.

Equation \ref{eqn:projp}, below shows the projection after exponential mapping $y$ on $\mathcal{T}_x\mathbb{H}^{d, K}$ to the hyperboloid manifold $\mathbb{H}^{d, K}$. In our work we use $K=1$ for applying projections.
\begin{equation}
    Proj(exp_x^{\mathbb{H}, K} (y))  = (\sqrt{1 + \lvert \lvert v_{1:d} \rvert \rvert_2 ^2}, v_{1:d})
    \label{eqn:projp}
\end{equation}

The projection when mapping a point $y$ on $\mathcal{T}_x\mathbb{B}^{d, K}$ to Poincare space $\mathbb{B}^{d, K}$ is achieved by normalising $exp_x^{\mathbb{B},K}(y)$ if $\lvert \lvert exp_x^{\mathbb{B},K}(y) \rvert \rvert _2 ^2 > 1/\sqrt{K}$ or $>1$ in the case of the Poincare unit ball.

\subsection{Training details and architectures}

\subsubsection{Discrete surrogate model}
We consider an image dimension of $32 \times 32$ and $64 \times 64$ for the MNIST and AFHQ dataset respectively. Our modulation layer reduces the number features (latent dimension) of $z$ from 64 to 8 channels  and 1024 to 32 channels for the MNIST and AFHQ experiments respectively.
We perform all hyperbolic linear operations in $\mathcal{T}_x\mathbb{H}$ and do not directly update the embeddings $\zeta^{i} \in \mathbb{H}^{1,d}$ or $\in \mathbb{B}^{1,d}$ which means we can perform Euclidean optimisation. Therefore, for the experiments in the main paper, we use Adam optimisation with a learning rate of 0.0002 and batch size of 50 in all experiments to train the discrete surrogate model for 40 epochs. We experimented with Riemmanian stochastic gradient descent \cite{zhang2016riemannian} but found Adam optimisation to be more stable. We also found that $\zeta^n \in \mathbb{H}^{M_i \times d,1}$ provided greater stability during training and showed generally improved performance (see codebook ablation tables \ref{table:hypabl} and \ref{table:poiabl} below in section \ref{subsec:cbls})
% Note, we do not perform linear operations in $\mathcal{T}_x\mathbb{B}$ due to stability issues. Also, note we do not directly update the embeddings of $\zeta^{i}$ in $\mathbb{H}^{1,d}$ or $\mathbb{B}^{1,d}$ as the  embeddings are updated only as a function of the learned hyperboloid linear operations in $\mathcal{T}_x\mathbb{H}$ followed by a mapping to $\mathcal{T}_x\mathbb{B}$. 

% Furthermore, it is important state, that low dimensional embeddings for $\zeta^0$, for example where $d = 2$ will have an adverse effect on the output of the decoder, producing very blurry images. In this case, one can use the Euclidean codebook with higher dimensionality $d$ for  and theof 
% The only embeddings we update directly through vector quantisation are in the the Euclidean codebook $\mathbb{E}\in \mathbb{R}^{d' \times M_0}$. \cite{chami2019hyperbolic} applies hyperboloid non-linear activation to aid with the smoothly varying curvature. We  do not include non-linearity in our hyperboloid linear layer which uses fixed negative curvature as it did not add any benefit to training convergence or knowledge distillation performance.

We described the optimisation of the Binary weights in the main paper. In our experiments, we use, adaptivity rate $\gamma$ of $0.0004$ and threshold  $\tau$ of $1e-8$. The adaptivity rate is analogous to the learning rate and can be seen as the consistency of a gradient signal required to induce a flip in weights from $0$ to $1$ or vice versa \cite{helwegen2019latent}. A high $\gamma$ can therefore induce a flip quicker given a new gradient signal but this can also mean noisy training \cite{helwegen2019latent}. $\tau$ is reflective of the strength of the gradient signal required to induce a flip \cite{helwegen2019latent}.

\subsubsection{Decoder}
The Decoder for MNIST experiments initially consists of a $1 \times 1$ convolutional block to increase the number of channels back to 64 channels before 3 up-sampling blocks. Each up-sampling block consists of bi-linear interpolation before two pre-activation $3 \times 3$ convolutional blocks with residual connections \cite{he2016identity}. We use batch normalisation and ReLU activation. 
In our AFHQ experiments, the decoder also initially starts with a $1 \times 1$ convolutional block to increase the number of channels back to 1024 before 4 up-sampling blocks identical to the one used in our MNIST experiments. We train the decoder in both experiments for 40 epochs using Adam optimisation with a learning rate of 0.0002.

\subsubsection{Pre-trained classifiers}

The MNIST pre-trained classifier consists of 7 convolutional blocks made up of a $3 \times 3$ convolutional layer followed by batch normalisation and ReLU non-linearity. This is followed by global average pooling and a single layer linear classifier. Max pooling is applied after the first, third and fifth layers. The number of channels corresponding to the final convolutional block is 64. We train this classifier for 50 epochs with a batch size of 50. We use Adam optimisation with an initial learning rate of 0.01 and weight decay of 0.01. We achieve $99\%$ accuracy with this pre-trained classifier

We use the DenseNet-121 as our pre-trained classifier for the AFHQ dataset \cite{iandola2014densenet}. We train using Adam optimisation with an initial learning rate of 0.001 and weight decay of 0.05. $98\%$ accuracy is achieved with this classifier. We reduce the dimensionality of the MNIST and AFHQ images to $4\times 4$ in the final layer before being inputted into the classifier. 

\subsubsection{Optional commitment loss}

Given $z_{pq}^{i} \in \mathbb{R}^{K\times d'}$ and $k \in K$, we define an optional commitment loss to move each $z_{pqk}^{i} \in z_{pq}^{i}$ closer to its sampled codebook vector denoted $\zeta^i_k$ shown below in Equation \ref{eqn:cl}.
\begin{equation}
       \mathcal{L}_{cb} = \sum_{i=0}^{i=n} \sum_{t=0}^{t=K}d^{\mathbb{B}, 1}(z_{pqk}^{i}), \ sg (\zeta^i_k)).
       \label{eqn:cl}
\end{equation}      

Therefore the total loss in this case is: $\mathcal{L}_{Total} = \mathcal{L}_{dist} + \mathcal{L}_{quant} + \mathcal{L}_{Poincare} + \epsilon \mathcal{L}_{cb}, \ \epsilon \in \{0,1\}$. 

We found this to not affect knowledge distillation performance except to provide a more committed notion of abstraction for $z_{q}^{i}$

\subsection{Codebook ablations}
\label{subsec:cbls}

We run multiple codebook ablations to find the minimum number of codebook vectors required to achieve a knowledge distillation accuracy of at least $90\%$ shown in table \ref{table:hypabl} and \ref{table:poiabl}. We fix the number of codebook vectors in $\zeta^n$ to $\lfloor log_2 N + 1\rfloor$. We note in more complex datasets, a larger number of codebook vectors maybe required in $\zeta^n$, whereby a class may need to be encoded with more than one possible combination of codebook vectors in order to fit the data. We also compare knowledge distillation accuracy of codebooks with Poincare embeddings against hyperboloid embeddings. In these experiments we only use a three level hierarchy. The user can however decide if they would like more levels of abstraction by increasing the number of codebooks. We note also, the use of codebook ablations is a limitation for the practicality of our explainability method.

\FloatBarrier
\begin{table}[H]
\vspace{-10pt}
\centering
\caption{Knowledge distillation accuracy of different codebook sizes for Poincare, Hyperboloid and Euclidean embeddings on the MNIST dataset. The best knowledge distillation accuracy for each embedding dimension is highlighted in bold.}
\vspace{10pt}
\label{table:hypabl}
\begin{tabular}{@{}cccccccccc@{}}
\toprule
\multicolumn{1}{c}{\textbf{Embedding dim. ($\rightarrow$)}}&
\multicolumn{3}{c}{\textbf{Poincare}}
& \multicolumn{3}{c}{\textbf{Hyperboloid}}
& \multicolumn{3}{c}{\textbf{Euclidean}}

\\ \cmidrule(l){2-10} 
\multicolumn{1}{c}{\textbf{Codebook Size ( $\zeta^0, \zeta^1,\zeta^2 $) ($\downarrow$)}}&
\multicolumn{1}{c}{\textbf{2}}         
& \multicolumn{1}{c}{\textbf{4}} &
\multicolumn{1}{c}{\textbf{16}} &        
\multicolumn{1}{c}{\textbf{2}}&         
\multicolumn{1}{c}{\textbf{4}} &
\multicolumn{1}{c}{\textbf{16}}&
\multicolumn{1}{c}{\textbf{2}}&         
\multicolumn{1}{c}{\textbf{4}} &
\multicolumn{1}{c}{\textbf{16}}
\\ \cmidrule(l){1-10}

\multicolumn{1}{c}{512, 64, 4} & \multicolumn{1}{c}{0.92} & \multicolumn{1}{c}{0.91} &
\multicolumn{1}{c}{0.98} & \multicolumn{1}{c}{\textbf{0.95}} &      \multicolumn{1}{c}{\textbf{0.93}} & \multicolumn{1}{c}{\textbf{0.99}} &
\multicolumn{1}{c}{0.80} &      \multicolumn{1}{c}{0.90} & \multicolumn{1}{c}{0.95} 
\\

\multicolumn{1}{c}{256, 64, 4} & \multicolumn{1}{c}{\textbf{0.94}} & \multicolumn{1}{c}{0.94} &
\multicolumn{1}{c}{\textbf{0.99}} & \multicolumn{1}{c}{0.91} &      \multicolumn{1}{c}{\textbf{0.98}} & \multicolumn{1}{c}{\textbf{0.99}} &
\multicolumn{1}{c}{0.86} &      \multicolumn{1}{c}{0.92} & \multicolumn{1}{c}{0.96}
\\

\multicolumn{1}{c}{256, 32, 4}& \multicolumn{1}{c}{0.89} & \multicolumn{1}{c}{0.96} &
\multicolumn{1}{c}{\textbf{0.99}} & \multicolumn{1}{c}{\textbf{0.93}} &      \multicolumn{1}{c}{\textbf{0.98}} & \multicolumn{1}{c}{0.97} &
\multicolumn{1}{c}{0.79} &      \multicolumn{1}{c}{0.91} & \multicolumn{1}{c}{0.98}
\\

\multicolumn{1}{c}{128, 32, 4}& \multicolumn{1}{c}{0.90} & \multicolumn{1}{c}{\textbf{0.96}} &
\multicolumn{1}{c}{\textbf{0.99}} & \multicolumn{1}{c}{\textbf{0.91}} &      \multicolumn{1}{c}{0.95} & \multicolumn{1}{c}{\textbf{0.99}} &
\multicolumn{1}{c}{0.81} &      \multicolumn{1}{c}{0.92} & \multicolumn{1}{c}{0.95}
\\

\multicolumn{1}{c}{64, 16, 4}& \multicolumn{1}{c}{0.85} & \multicolumn{1}{c}{\textbf{0.88}} &
\multicolumn{1}{c}{\textbf{0.94}} & \multicolumn{1}{c}{\textbf{0.90}} &      \multicolumn{1}{c}{\textbf{0.88}} & \multicolumn{1}{c}{0.93} &
\multicolumn{1}{c}{0.80} &      \multicolumn{1}{c}{0.83} & \multicolumn{1}{c}{0.90}
\\ \bottomrule
\end{tabular}
\end{table}
\FloatBarrier
\FloatBarrier

\begin{table}[H]
\vspace{-10pt}
\centering
\caption{Knowledge distillation accuracy of different codebook sizes for Poincare, Hyperboloid and Euclidean embeddings on the AFHQ dataset. The best knowledge distillation accuracy for each embedding dimension is highlighted in bold.}
\vspace{10pt}

\label{table:poiabl}
\begin{tabular}{@{}cccccccccc@{}}
\toprule
\multicolumn{1}{c}{\textbf{Embedding dim. ($\rightarrow$)}}&
\multicolumn{3}{c}{\textbf{Poincare}}
& \multicolumn{3}{c}{\textbf{Hyperboloid}}
& \multicolumn{3}{c}{\textbf{Euclidean}}

\\ \cmidrule(l){2-10} 
\multicolumn{1}{c}{\textbf{Codebook Size ( $\zeta^0, \zeta^1,\zeta^2 $) ($\downarrow$)}}&
\multicolumn{1}{c}{\textbf{2}}         
& \multicolumn{1}{c}{\textbf{4}} &
\multicolumn{1}{c}{\textbf{16}} &        
\multicolumn{1}{c}{\textbf{2}}&         
\multicolumn{1}{c}{\textbf{4}} &
\multicolumn{1}{c}{\textbf{16}}&
\multicolumn{1}{c}{\textbf{2}}&         
\multicolumn{1}{c}{\textbf{4}} &
\multicolumn{1}{c}{\textbf{16}}
\\ \cmidrule(l){1-10}

\multicolumn{1}{c}{512, 64, 2}& \multicolumn{1}{c}{\textbf{0.94}} & \multicolumn{1}{c}{0.92} &
\multicolumn{1}{c}{0.96} & \multicolumn{1}{c}{0.90} &      \multicolumn{1}{c}{\textbf{0.95}} & \multicolumn{1}{c}{\textbf{0.98}} &
\multicolumn{1}{c}{0.83} &      \multicolumn{1}{c}{0.91} & \multicolumn{1}{c}{0.93} 
\\

\multicolumn{1}{c}{256, 64, 2}& \multicolumn{1}{c}{0.90} & \multicolumn{1}{c}{0.95} &
\multicolumn{1}{c}{0.98} & \multicolumn{1}{c}{\textbf{0.92}} &      \multicolumn{1}{c}{\textbf{0.98}} & \multicolumn{1}{c}{\textbf{0.99}} &
\multicolumn{1}{c}{0.80} &      \multicolumn{1}{c}{0.90} & \multicolumn{1}{c}{0.97}
\\

\multicolumn{1}{c}{256, 32, 2}& \multicolumn{1}{c}{0.86} & \multicolumn{1}{c}{0.85} &
\multicolumn{1}{c}{0.93} & \multicolumn{1}{c}{\textbf{0.88}} &      \multicolumn{1}{c}{\textbf{0.91}} & \multicolumn{1}{c}{\textbf{0.96}} &
\multicolumn{1}{c}{0.77} &      \multicolumn{1}{c}{0.81} & \multicolumn{1}{c}{0.88}
\\

\multicolumn{1}{c}{128, 32, 2}& \multicolumn{1}{c}{\textbf{0.88}} & \multicolumn{1}{c}{0.88} &
\multicolumn{1}{c}{0.91} & \multicolumn{1}{c}{0.84} &      \multicolumn{1}{c}{\textbf{0.92}} & \multicolumn{1}{c}{\textbf{0.96}} &
\multicolumn{1}{c}{0.80} &      \multicolumn{1}{c}{0.86} & \multicolumn{1}{c}{0.90}
\\ 

\multicolumn{1}{c}{64, 16, 2}& \multicolumn{1}{c}{0.78} & \multicolumn{1}{c}{0.85} &
\multicolumn{1}{c}{0.83} & \multicolumn{1}{c}{\textbf{0.81}} &      \multicolumn{1}{c}{\textbf{0.90}} & \multicolumn{1}{c}{\textbf{0.92}} &
\multicolumn{1}{c}{0.75} &      \multicolumn{1}{c}{0.80} & \multicolumn{1}{c}{0.84}
\\ \bottomrule

\end{tabular}
\end{table}
\FloatBarrier

\subsection{Additional results}
\begin{figure}[hbt!]
    \centering
    \subfloat[Image-level tree]{\includegraphics[width=.3\textwidth]{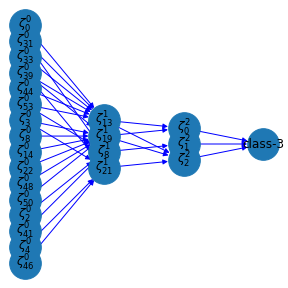}} \hfill
    \subfloat[Visual rules]{\includegraphics[width=.6\textwidth]{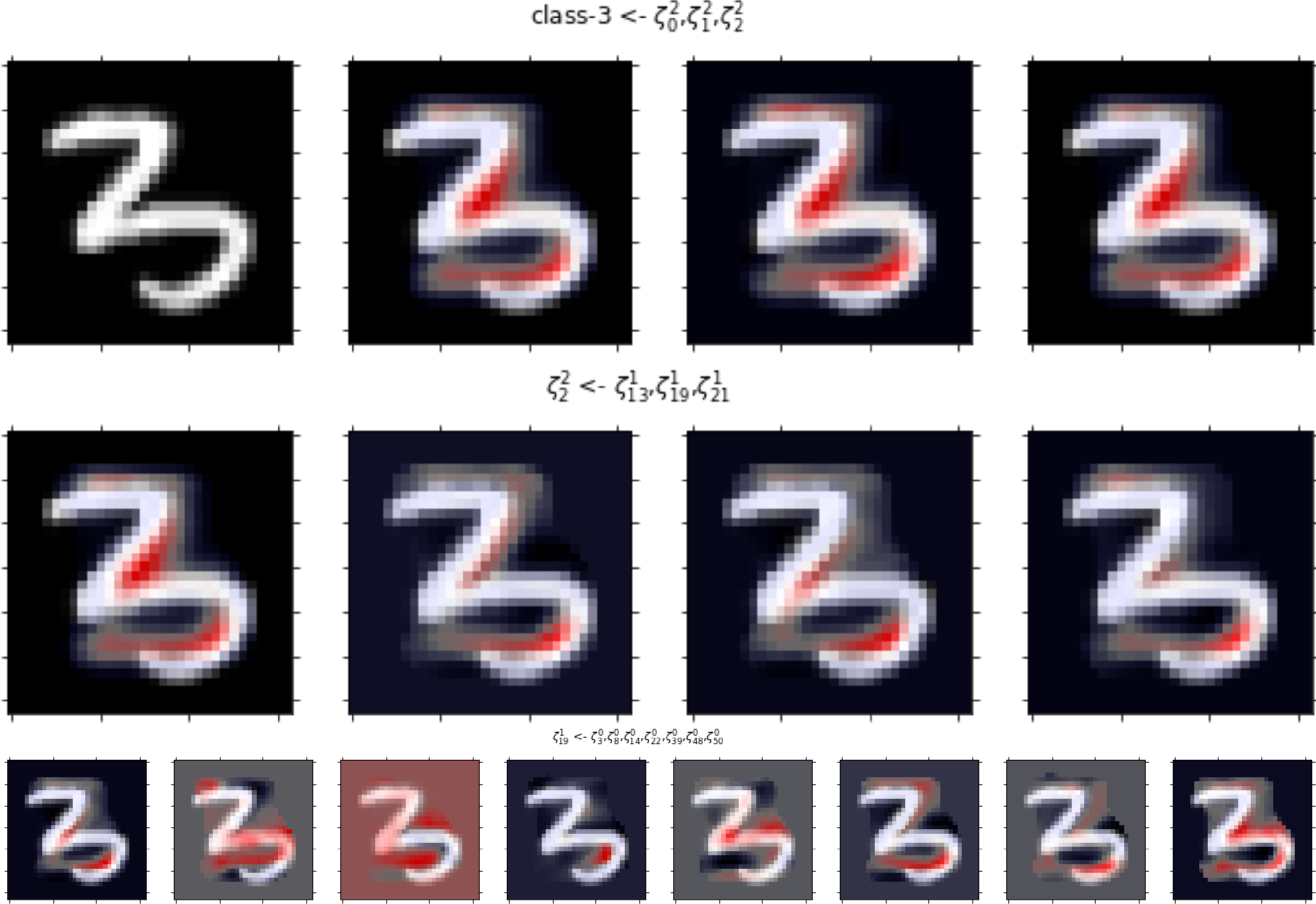}} \hfill
    \caption{This figure describes the explanations obtained using our proposed framework on the MNIST classifier for class '3', with hierarchical rules: $class3(x) \leftarrow \zeta^2_0(x), \zeta^2_1(x),\zeta^2_2(x)$, where $\zeta^2_2(x) \leftarrow \zeta^1_{13}(x),\zeta^1_{19}(x),\zeta^1_{21}(x)$, and $\zeta^1_{19}(x) \leftarrow \zeta^0_3(x),\zeta^0_8(x),\zeta^0_{14}(x),\zeta^0_{22}(x),\zeta^0_{48}(x),\zeta^0_{50}(x)$ }
    \label{fig:mnistexp3}
    \vspace{-5pt}
\end{figure}

\begin{figure}[hbt!]
    \centering
    \subfloat[Image-level tree]{\includegraphics[width=.3\textwidth]{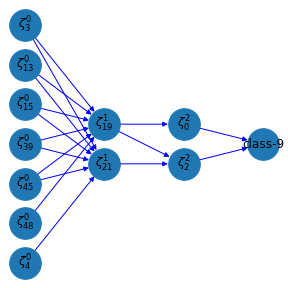}} \hfill
    \subfloat[Visual rules]{\includegraphics[width=.6\textwidth]{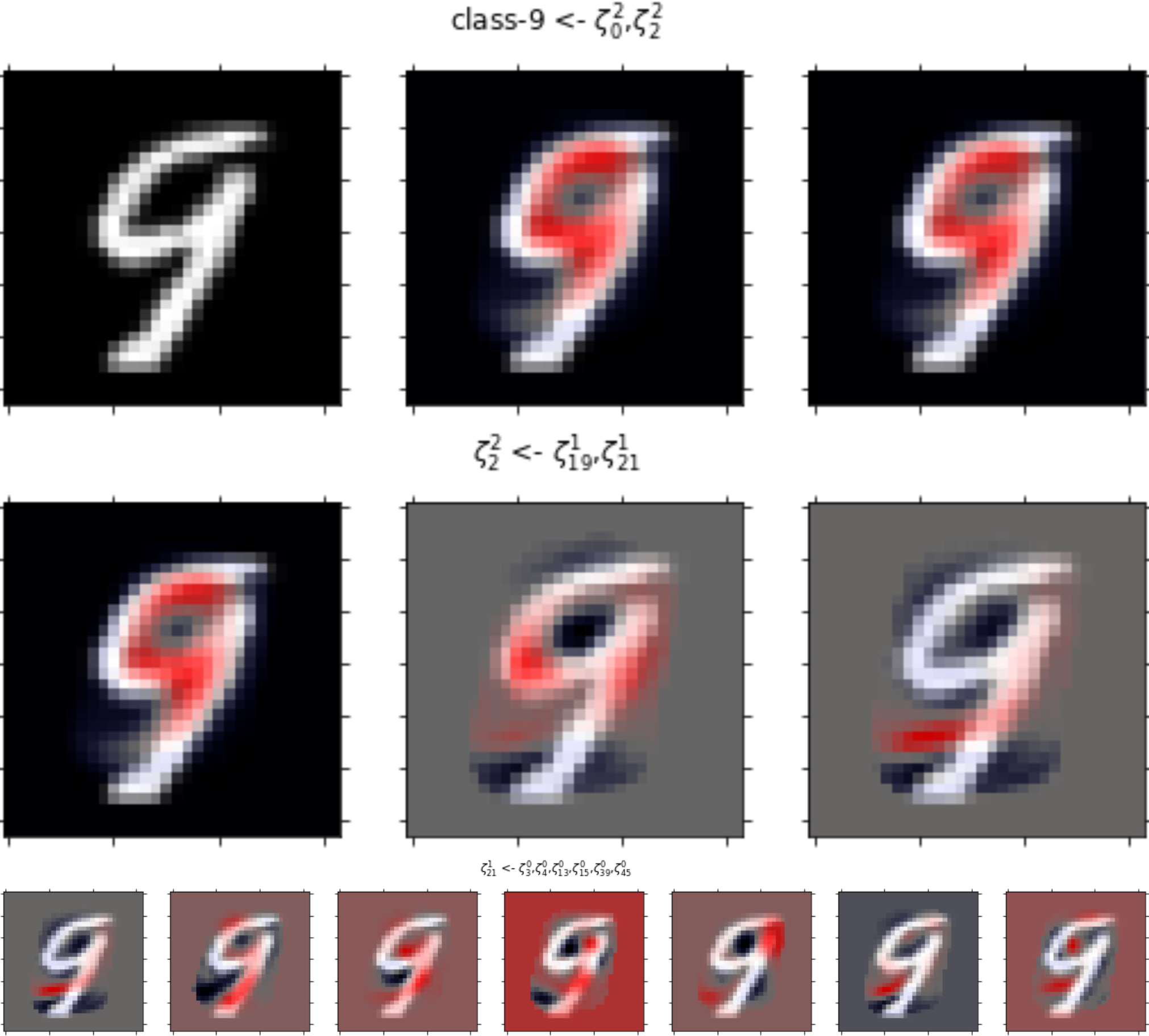}} \hfill
    \caption{This figure describes the explanations obtained using proposed framework on the MNIST classifier for class '9', with hierarchical rules: $class9(x) \leftarrow \zeta^2_0(x), \zeta^2_2(x)$, where $\zeta^2_2(x) \leftarrow \zeta^1_{19}(x),\zeta^1_{21}(x)$, and $\zeta^1_{21}(x) \leftarrow \zeta^0_3(x),\zeta^0_4(x),\zeta^0_{13}(x),\zeta^0_{15}(x),\zeta^0_{45}(x)$}
    \label{fig:mnistexp4}
    \vspace{-5pt}
\end{figure}

\begin{figure}[hbt!]
    \centering
    \subfloat[Image-level tree]{\includegraphics[width=.3\textwidth]{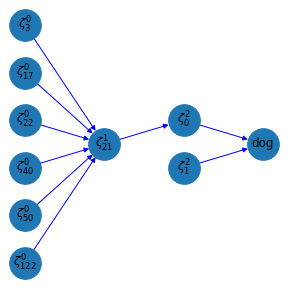}} \hfill
    \subfloat[Visual rules]{\includegraphics[width=.5\textwidth]{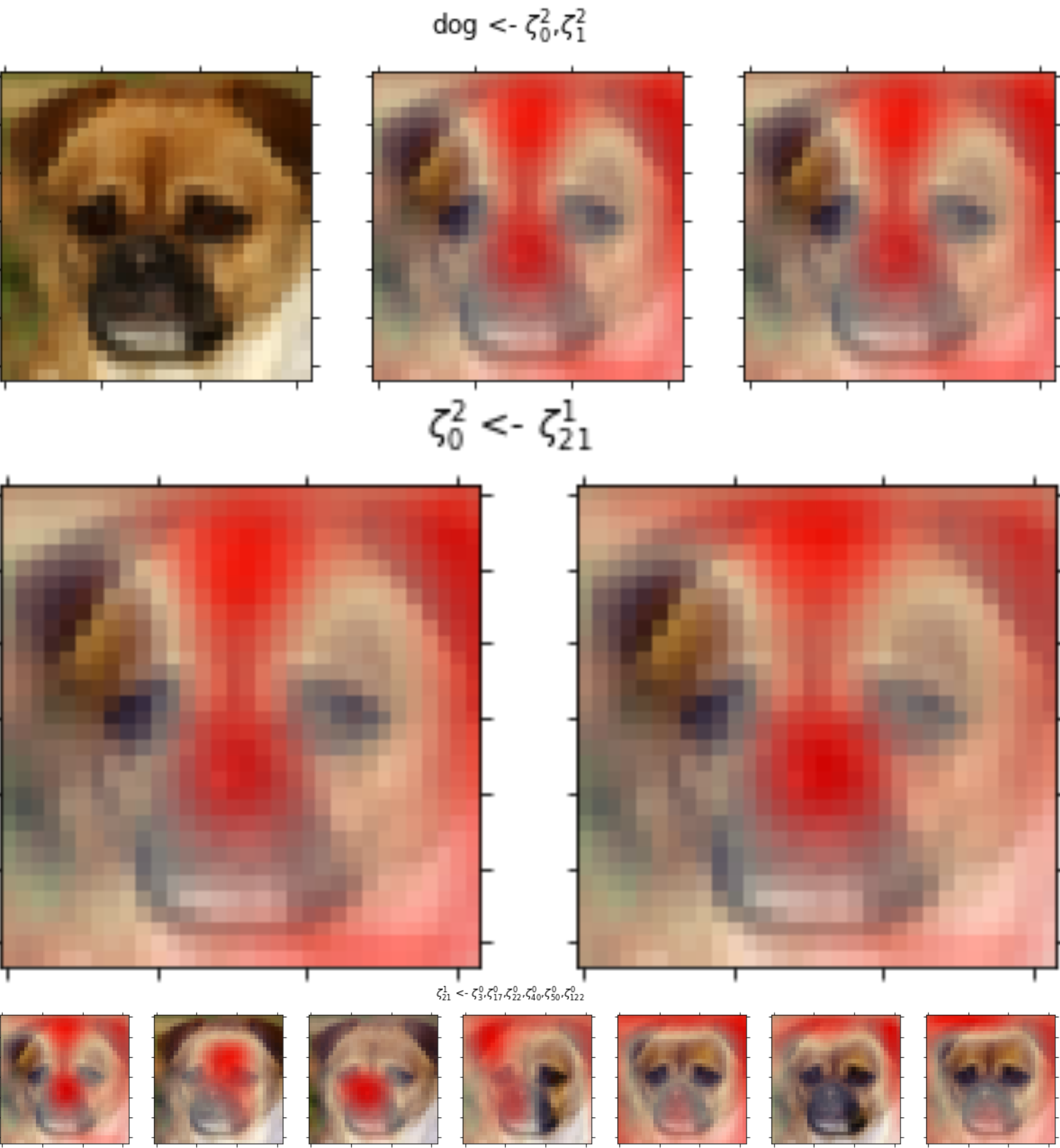}} \hfill
    \caption{This figure describes the explanations obtained using our proposed framework for the AFHQ classifier for class dog, with hierarchical rules: $dog(x) \leftarrow \zeta^2_0(x), \zeta^2_1(x)$, where $\zeta^2_0(x) \leftarrow \zeta^1_{21}(x)$, and $\zeta^1_{21}(x) \leftarrow \zeta^0_3(x),\zeta^0_{17}(x),\zeta^0_{22}(x),\zeta^0_{40}(x),\zeta^0_{122}(x),\zeta^0_{50}(x)$}
    \label{fig:afhqexp4}
    \vspace{-5pt}
\end{figure}

In this section, we provide additional examples illustrating the explanations of the proposed framework. 
Figure \ref{fig:mnistexp3} and \ref{fig:mnistexp4} describes our explanations for a model trained on the MNIST dataset classifying class '3', and '9' respectively.
Figure \ref{fig:afhqexp} and \ref{fig:afhqexp4} describes our explanations for a model trained on the AFHQ dataset classifying class 'dog' and 'cat' respectively.
The sub-figure (a) in Figure \ref{fig:mnistexp3}, \ref{fig:mnistexp4}, \ref{fig:afhqexp}, and \ref{fig:afhqexp4} demonstrates the local image-level abstraction tree reflecting how definite symbols from codebook $\zeta^0$ combine to form abstract symbols in a class, while the sub-figure (b) illustrates these abstractions with visual rules.

\begin{figure}[hbt!]
    \centering
    \includegraphics[width=0.5\textwidth]{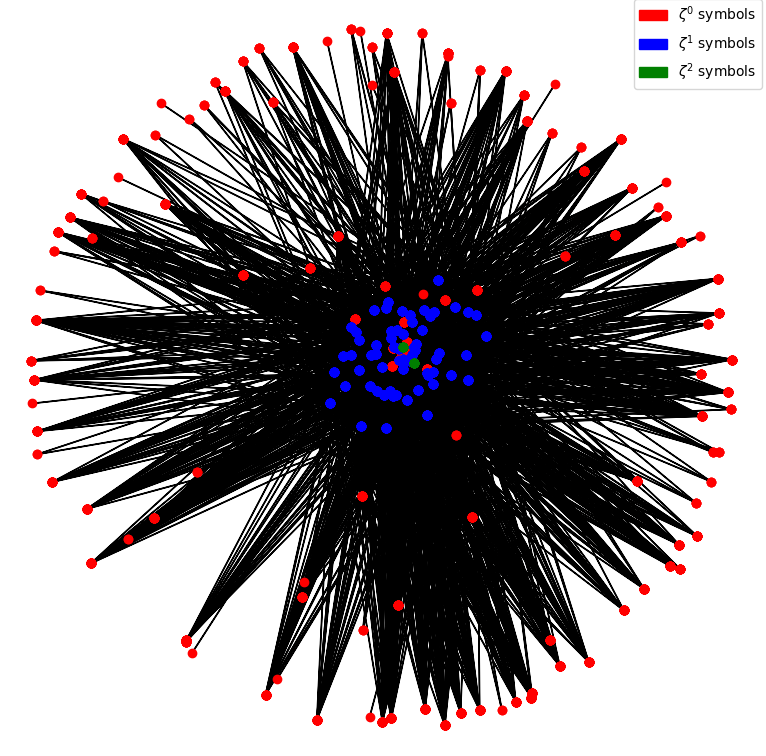}
    \caption{Poincare embedding of discrete symbols obtained for the AFHQ classifier. Here, red, blue, and green nodes indicate symbols from $\zeta^0, \zeta^1, \zeta^2$ layers abstraction.}
    \label{fig:poincare_afhq}
\end{figure}

\begin{figure}[hbt!]
    \centering
    \subfloat[Cat abstraction tree]{\includegraphics[width=.5\textwidth]{imgs/AFHQ-class-0-tree.png}} \hfill
    \subfloat[Dog abstraction tree]{\includegraphics[width=.5\textwidth]{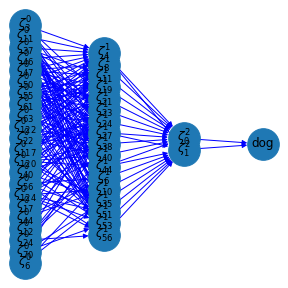}} \hfill
    \caption{This figure describes the difference between class-level trees for 'cat' and 'dog' class, illustrating the difference in symbols sampled from $\zeta^0, \zeta^1, \& \zeta^2$ in abstracting a particular class.}
    \label{fig:afhqexp2}
    \vspace{-5pt}
\end{figure}

Figure \ref{fig:poincare_afhq} demonstrates the distribution of codebook symbols on a Poincare disk. The symbols from $\zeta^0$ are spread along the circumference of the disk, while the symbols from $\zeta^1$ and $\zeta^2$ are distributed inside the disk, maintaining the hierarchy.
Finally, Figure \ref{fig:afhqexp2} demonstrates the class-level abstraction tree for class `cat' and `dog' in the AFHQ dataset. We can observe distinctions between symbols sampled and abstracted to form the class trees. 

\begin{figure}[hbt!]
    \centering
    \subfloat[Image-level tree]{\includegraphics[width=.3\textwidth]{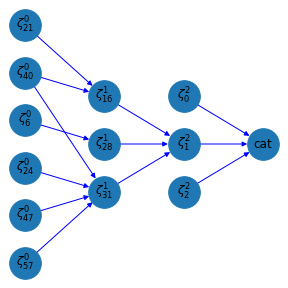}} \hfill
    \subfloat[Visual rules]{\includegraphics[width=.6\textwidth]{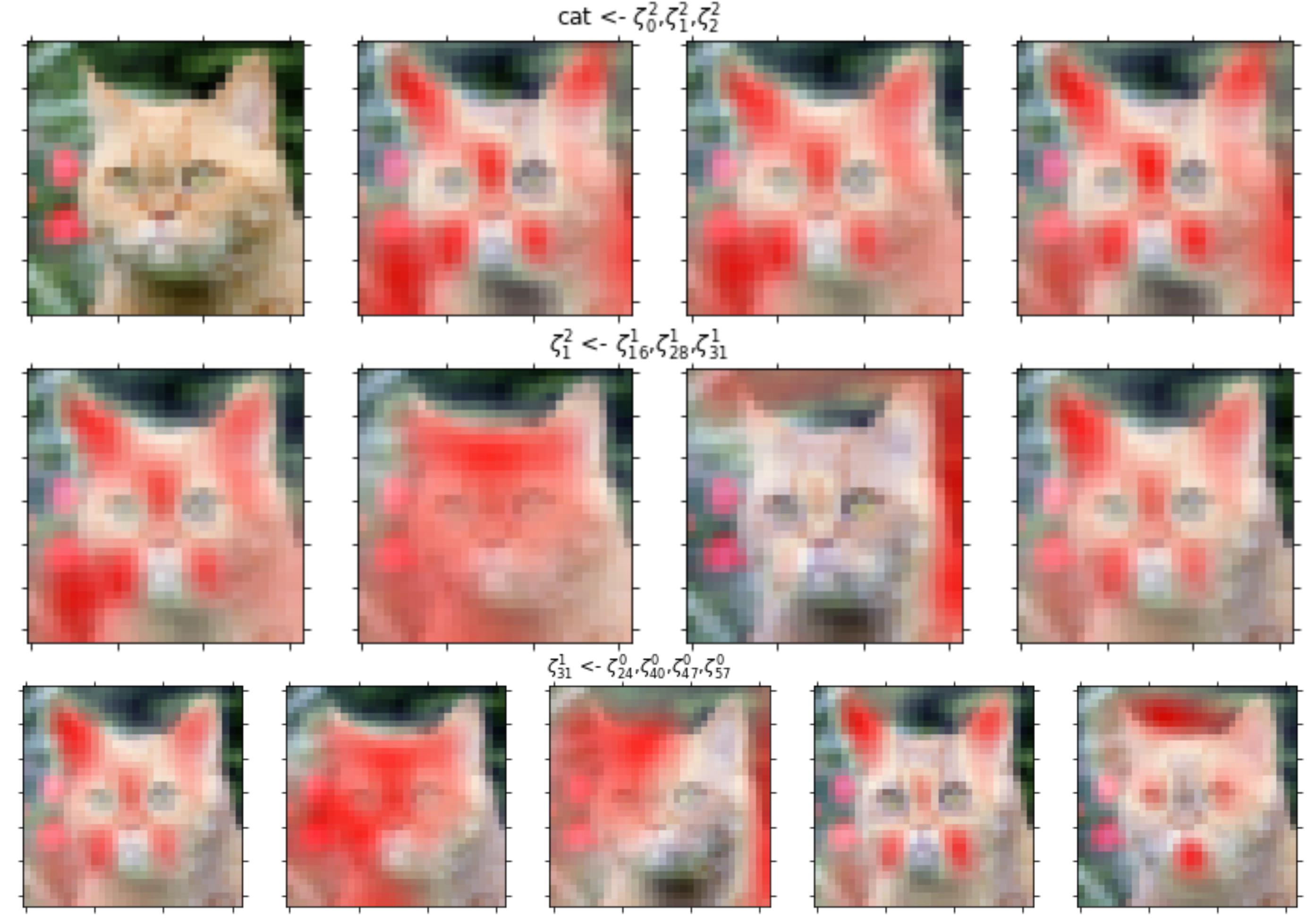}} \hfill
    \caption{This figure describes the explanations obtained using our proposed framework for the AFHQ classifier classifying class 'cat' (ablation with 3 symbols in the final codebook of the hierarchy), with hierarchical rules: $cat(x) \leftarrow \zeta^2_0(x), \zeta^2_1(x),\zeta^2_2(x)$, where $\zeta^2_1(x) \leftarrow \zeta^1_{16}(x),\zeta^1_{31}(x),\zeta^1_{28}(x)$, and $\zeta^1_{31}(x) \leftarrow \zeta^0_{24}(x),\zeta^0_{40}(x),\zeta^0_{57}(x),\zeta^0_{47}(x)$.}
    \label{fig:afhqexp}
    \vspace{-5pt}
\end{figure} 

\end{document}